%% file: main_arxiv.tex
\documentclass{article} %
\usepackage[dvipsnames,table]{xcolor}         %
\usepackage{iclr2027_conference,times}

\usepackage[utf8]{inputenc} %
\usepackage[T1]{fontenc}    %
\usepackage[pagebackref=true,breaklinks=true,colorlinks,citecolor=gray,linkcolor=RoyalBlue]{hyperref}
\usepackage{url}            %
\usepackage{booktabs}       %
\usepackage{amsfonts}       %
\usepackage{nicefrac}       %
\usepackage{microtype}      %
\usepackage{microtype}
\usepackage{graphicx}
\usepackage{subfigure}
\usepackage{booktabs} %
\usepackage{natbib}
\usepackage{lipsum}

\usepackage{xcolor}
\definecolor{mypink}{HTML}{DE23EB}
\newcommand{\pinkbox}{\textcolor{mypink!55!white}{\rule{0.7em}{0.7em}}}

\usepackage{algorithm}
\usepackage{algpseudocode}
\let\STATE\State
\algrenewcommand{\algorithmicindent}{0.7em}

\usepackage{amsmath}
\usepackage{amssymb}
\usepackage{mathtools}
\usepackage{amsthm}
\usepackage{float}
\usepackage{xcolor}
\usepackage{xspace}
\usepackage{makecell}
\usepackage{wrapfig}
\usepackage{pifont}%
\usepackage{caption}
\DeclareCaptionFont{customsize}{\fontsize{9.5}{10.5}\selectfont}
\usepackage{amsthm}

\usepackage{thmtools,thm-restate}

\usepackage[capitalize,noabbrev]{cleveref}

\newcommand{\ctxenc}{\ensuremath{g_\phi}\xspace}
\newcommand{\zp}{\ensuremath{\mathbf{z}_{\text{phy}}}\xspace}
\newcommand{\encv}{\ensuremath{E_{\text{v}}}\xspace}

\definecolor{highlight}{rgb}{0,0,0}

\newcommand{\method}{PhysiCalWorld\xspace}
\newcommand{\shortmethod}{PhysiCal\xspace}

\usepackage{makecell}
\usepackage{multirow}
\usepackage{multicol}
\usepackage{setspace}
\usepackage[export]{adjustbox}

\usepackage{enumitem}

\input{math_commands.tex}

\usepackage{hyperref}
\usepackage{url}

\definecolor{commentgray}{RGB}{119,119,119}

\definecolor{salmon}{rgb}{1.0, 0.55, 0.41}
\definecolor{darksalmon}{rgb}{0.91, 0.59, 0.48}
\definecolor{darkpink}{rgb}{0.91, 0.33, 0.5}
\newcommand{\commentcode}[1]{\textcolor{commentgray}{\# #1}}

\newcommand{\hnam}[1]{{\textcolor{darkpink}{hnam}}}

\newcommand{\vspacet}[1]{\vspace{#1}}

\usepackage{titlesec}

\titlespacing*{\paragraph}
  {0pt}      %
  {0pt}      %
  {4pt}      %

\newcommand{\pcworld}{{\fontfamily{CrimsonPro-TLF}\selectfont PhysiCalWorld}}

\makeatletter
\newcommand{\printfnsymbol}[1]{%
  \textsuperscript{\@fnsymbol{#1}}%
}
\makeatother

\title{ \pcworld: \hspace{-3pt}In-Context Physics Calibration for World Models}

\newcommand{\asep}{\hspace{1.2em}}  %

\author{
Yunhao Luo\printfnsymbol{1}\asep Wentao He\printfnsymbol{1}\asep Yinpei Dai\asep Hyelin Nam\asep Hongze Fu \\
~\textbf{Joyce Chai\asep Nima Fazeli\printfnsymbol{2}\asep Jeong Joon Park\printfnsymbol{2}} \\[4pt]
~University of Michigan
}

\iclrfinalcopy
\begin{document}

\maketitle

{\renewcommand{\thefootnote}{\fnsymbol{footnote}}
\footnotetext{\printfnsymbol{1}Equal contribution. \quad \printfnsymbol{2}Equal advising.}
}

\begin{abstract}

\input{sections/abstract}

\end{abstract}

\input{sections/introduction}

\input{sections/related_work}

\input{sections/method}

\input{sections/experiments}

\input{sections/discussion}

\input{sections/acknowledgement}

\bibliography{iclr2027_conference}
\bibliographystyle{iclr2027_conference}

\appendix

\input{sections/appendix}

\end{document}

%% file: math_commands.tex
\usepackage{amsmath,amsfonts,bm}

\def\eqref#1{equation~\ref{#1}}

\def\1{\bm{1}}

\DeclareMathAlphabet{\mathsfit}{\encodingdefault}{\sfdefault}{m}{sl}
\SetMathAlphabet{\mathsfit}{bold}{\encodingdefault}{\sfdefault}{bx}{n}

%% file: sections/abstract.tex
Existing world models typically perceive a scene through current visual frame or a narrow history window. However, these observations may not reveal the physics properties that determine transition dynamics, such as an object's mass distribution, joint damping, or surface friction. Without calibrating these physics factors for world models, predictions tend to regress toward an empirical prior and induce severe drift in novel test scenes.
To this end, we propose \method, which calibrates action-conditioned world models to the instance-specific dynamics via in-context prompting, without fine-tuning or analytical physics modeling. Our method encodes an exploratory interaction clip collected within the test scene into a compact physics descriptor token, which grounds subsequent predictions in the scene's actual physical behavior.
\method is trained end-to-end with the latent prediction loss alone, without physics labels. At inference time, the descriptor token is extracted once and can be reused across rollouts with negligible compute overhead.
We evaluate \method across four environments covering diverse physics variations and show that it substantially improves rollout prediction accuracy and visual planning success rate.
The learned descriptors also form a physically structured latent space and generalize beyond the physics ranges seen during training.
Project page at \href{https://physical-world.github.io/}{https://physical-world.github.io/}.

%% file: sections/introduction.tex
\section{Introduction}

World models predict how a scene evolves given its current state and
the intended actions of the agents within it. 
Trained on internet-scale data, video-based world
models \citep{bruce2024genie,yang2023learning,team2026advancing,agarwal2026cosmos}
are now able to generate photorealistic visual predictions of complex scene dynamics that are difficult to model with traditional physics engines.
Modeling the world in image space is thus becoming a promising alternative to hand-built physics modeling in such areas as robotics \citep{hou2026world} and autonomous driving \citep{feng2025survey}.

Existing world models usually generate predictions from the current frame or a short visual history. However, such observations may leave the underlying physical properties unconstrained, forcing the world models to guess a future state among multiple possible ones. For example, an opaque bottle appears identical whether or not it is filled with water, yet it will respond differently under identical actuation. A world model that relies only on empirical priors over object category or appearance will produce incorrect predictions in such scenarios, limiting its applicability in downstream applications that require accurate dynamics~\citep{zhou2025dino,quevedo2026worldgym}. This motivates \emph{calibrating} the world model to the hidden physical properties of each scene.

A natural way to resolve this physics ambiguity is to physically interact with the scene to infer its hidden properties, similar to how we humans would do for an object with unknown properties. 
Inspired by this, 
prior system identification methods~\citep{wu2015galileo,yu2017preparing,kumar2021rma} attempt to construct accurate physical models with closed-form equations and fit the predefined physics parameters to the in-situ interaction data. However, real-world systems exhibit complex, coupled physics factors, thus building a geometric model or formulating closed-form equations can be challenging.
Hence, to adapt to each deployment setting, existing world models~\citep{gao2026dreamdojo} are typically finetuned on a small set of in-house data, which is computationally expensive.

To this end, we propose \method, a method for in-context \textbf{Physi}cs \textbf{Cal}ibration for \textbf{World} models.  
Inspired by the in-context learning capabilities of large sequence models~\citep{brown2020language,bar2022visual}, we prompt world models with a sequence of observation-action pairs 
collected in the test environment (\textit{i.e.,} an environment whose physics configuration is unknown) for physics calibration. 
Our in-context prompting method sidesteps both the need for human effort to build analytical physics simulators for system identification and expensive model finetuning. Our method requires only an exploratory interaction clip without any gradient descent steps.
\method compresses the interaction sequence into a compact \emph{physics descriptor token} that conditions the world model. The entire system is trained end-to-end with only a video prediction objective, requiring neither ground-truth physics parameters nor a simulator.

At test time, \method first extracts the descriptor from a few exploratory interactions with the test scene (e.g., via random exploration or scripted policies). 
This descriptor then conditions all subsequent rollouts in that scene, enabling the world model to generate physically consistent predictions in novel configurations without any test-time weight updates.

In summary, our contributions are three-fold:
\vspace{-1pt}
\begin{itemize}[
  leftmargin=2em,   %
  topsep=0pt,       %
  itemsep=2pt,      %
  parsep=0pt,       %
  partopsep=0pt     %
]
\item \textbf{In-Context Physics Calibration:} We formulate physics calibration of generative world models as in-context conditioning: from a short interaction sequence, the model adapts zero-shot to instance-specific dynamics, with no test-time weight updates.
\item \textbf{Physics Descriptors Learned from Video Alone:} We introduce a \emph{physics descriptor token} that grounds the world model in hidden physical properties. It is learned end-to-end from video prediction alone, requiring neither ground-truth physics parameters nor a simulator.

\item \textbf{Performance and Analysis:} Across four environments with single and coupled physics variations, \method\ consistently improves rollout accuracy and planning success on two world model backbones. Its single-token descriptor matches or exceeds raw-context conditioning at a fraction of the cost. Without physics labels, the descriptors form a physically structured latent space and extrapolate beyond the training parameter range.

\end{itemize}

\begin{figure}[t]
    \centering
    \begin{minipage}[t]{0.535\textwidth}
        \vspace{0pt}
        \centering
        \includegraphics[width=\linewidth]{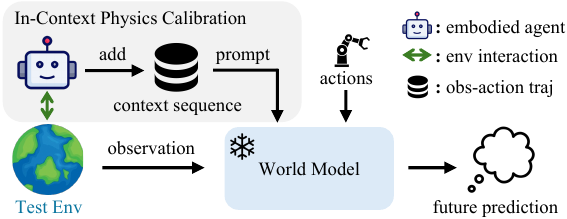}
    \end{minipage}\hfill
    \begin{minipage}[t]{0.45\textwidth}
        \vspace{3pt}
        \caption{  \textbf{In-Context Physics Calibration for World Models.}
        In a test environment with unknown transition dynamics, an embodied agent first executes exploratory actions to collect an in-context interaction sequence. This  sequence is then provided as prompt to calibrate the world model such that future predictions adhere to the actual dynamics of the test environment.
        }
        \label{fig:1_teaser}
    \end{minipage}
    \vspacet{-18pt}
\end{figure}

%% file: sections/related_work.tex
\section{Related Work}

\paragraph{World Models for Robotics.}
World models predict future states conditioned on current observations and agent actions, serving as an increasingly foundational framework for robotics~\citep{li2025comprehensive,hou2026world}. For example, they are scalable neural simulators, enabling low-cost policy evaluation without real-world execution~\citep{guo2026ctrl,gao2026dreamdojo,wang2026interactive,team2025evaluating,tseng2025scalable}. 
Through future imagination, world models can also substantially improve sample efficiency in reinforcement learning~\citep{hafner2019learning,hansen2022temporal,seo2023masked,wu2023daydreamer} and facilitate policy learning~\citep{saxena2026nostra,luo2025grounding}. 
Moreover, optimizing action sequences with world models enables embodied agents to complete novel tasks via visual trajectory planning~\citep{zhou2025dino,assran2025v,wang2024driving}. 
However, existing methods typically condition only on the current frame or a narrow observation history, assuming that these observations suffice to identify the underlying transition dynamics of the test scene.
As a result, predictions tend to suffer from severe drift if latent physical properties vary across test scenes.
In this work, we ground model predictions in the actual test-scene dynamics by introducing an additional in-context interaction prompt collected within the test scene as input to the world model, substantially enhancing the prediction accuracy and visual planning performance.

\paragraph{In-Context Learning in Physical AI.}
Context can serve embodied agents in various forms. For example, it can adapt a robot policy to test-scene dynamics by conditioning directly on execution history or by explicitly recovering physical parameters~\citep{lee2020learning,kumar2021rma,liu2025locoformer,zhang2025dynamics}. Another role is in-context policy learning, where task demonstrations steer actions either through structured representations such as keypoints and graphs~\citep{di2024keypoint, vosylius2025instant, hwang2024promptable} or raw video-action pairs~\citep{fu2024context, shah2025mimicdroid, ding2026contextflow},
and in-context reinforcement learning~\citep{lee2023supervised, lu2023structured}.
Most recently, context has been applied to action-conditioned world models by prepending context prefixes to adapt to new camera viewpoints or embodiments~\citep{yang2026syncworld}. 
However, prior works focus on policy adaptation, require privileged physical priors, or prepend raw context frames at high computational cost. In contrast, we use context to enable physics-grounded future prediction in world models, introducing a compact physics descriptor token to minimize computational overhead while maintaining high prediction accuracy and rollout efficiency.

\paragraph{Physics-Aware Video Generation.}
Video generative models have advanced rapidly~\citep{yang2025cogvideox,zheng2025open,chen2026sana}, with widespread commercial deployment~\citep{seedance2026seedance,wan2025wan}.
However, their adherence to physical laws remains  limited~\citep{motamed2026generative,upadhyay2026worldbench,bansal2025videophy,meng2024towards}.
Recent efforts seek to improve physical awareness in video models by incorporating explicit physics equations or parameters in the generation process~\citep{zhang2024physdreamer,wang2026physctrl,yuan2026newtongen,wang2026wisa},
coupling with external physics simulators~\citep{liu2024physgen,xie2025physanimator,foo2026physical}, 
leveraging foundation models~\citep{xue2025phyt2v,zhang2026videorepa}, or optimizing handcrafted rewards~\citep{zhang2026physrvg,li2025pisa}.
However, these approaches either rely on analytical formulations and are restricted to a pre-defined set of physics properties, or focus on learning general physics and tend to fail in scenarios that require fine-grained instance-level physics grounding.
In comparison, our work focuses on calibrating action-conditioned world models to the instance-specific dynamics of a test scene from an exploratory interaction prompt, operating entirely end-to-end without prior assumptions or ground-truth supervision of the environment's underlying physics.

%% file: sections/method.tex
\section{Method}
We consider an embodied agent interacting with an environment over discrete timesteps $t$. At each step, the agent receives a high-dimensional visual observation $o_t \in \mathcal{O} \subset \mathbb{R}^{H \times W \times 3}$ (and an optional proprioceptive state $p_t \in \mathbb{R}^{d_p}$) and executes a control action $a_t \in \mathcal{A} \subset \mathbb{R}^{d_a}$. Our objective is to learn an action-conditioned world model that predicts future observations directly in a compact latent space and enables both long-horizon rollout and visual planning.

\paragraph{Latent World Models.} 
While our proposed in-context prompting framework is a plug-and-play module applicable to different world model architectures, we base the proposed method on two recent latent world models, DINO-WM~\citep{zhou2025dino} and LeWM~\citep{maes2026leworldmodel}, due to their efficiency and strong trajectory planning performance.
In these latent world models,  each input RGB frame $o_t$ is encoded by a visual encoder \encv into a set of visual tokens $z_t$:
\begin{equation}z_t = \encv(o_t) \in \mathbb{R}^{N_z \times d_z},\end{equation}
where $N_z$ denotes the number of visual tokens per frame and $d_z$ is the feature dimension. 
Meanwhile, actions $a_t$ are mapped into action embeddings $e_t^a \in \mathbb{R}^{d_a'}$ via a learned MLP $E_\text{a}$:
\begin{equation}
e_t^a = E_\text{a}(a_t).
\end{equation}
The latent dynamics predictor $f_\theta$ (a causal transformer with weights $\theta$) models temporal transitions over a history window of $K$ steps. At step $t$, the predictor takes a sequence of past visual tokens and action embeddings to predict the next latent tokens:
\begin{equation}
\hat{z}_{t+1} = f_\theta\left(\{z_i, e_i^a\}_{i=t-K+1}^t\right).
\end{equation}
For visual evaluation and pixel-space rollouts, we train a lightweight convolutional decoder $D_\text{v}$ that reconstructs images $\hat{o}_t$ from latent tokens, $\hat{o}_{t+1} = D_\text{v}(\hat{z}_{t+1})$, while downstream visual planning can be conducted directly within the latent space $\mathcal{Z}$.

\begin{figure}[t]
  \centering
  \includegraphics[width=1\linewidth]{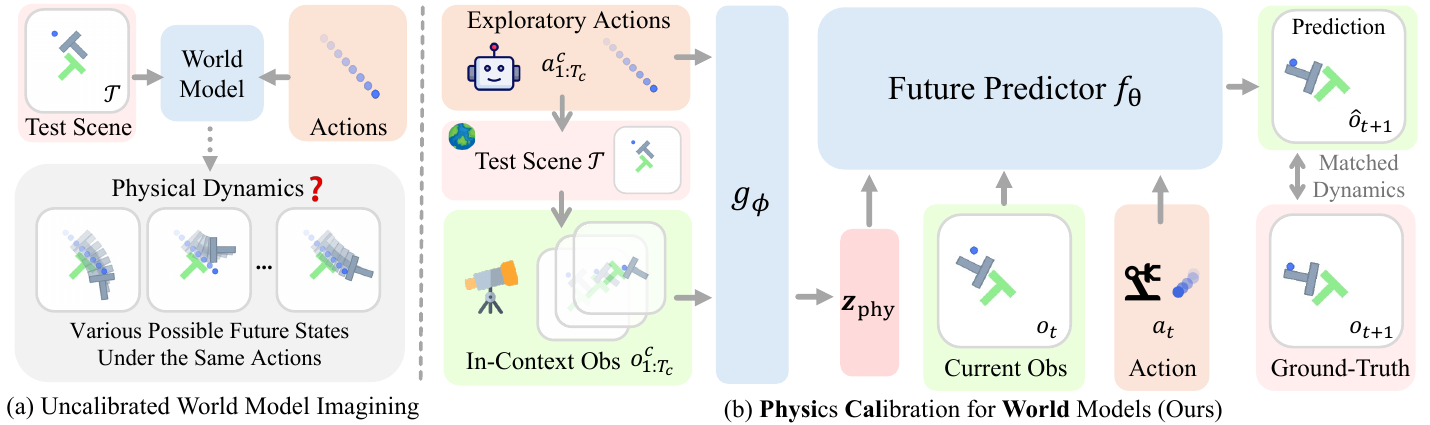}
  \vspacet{-16pt}
  \caption{ 
  \textbf{Method Overview.} 
  \method calibrates action-conditioned world models to the actual dynamics of a test scene via prompting with in-context interaction. 
  In \textbf{(a)}, the current visual observation alone cannot identify the physical dynamics of the test scene $\mathcal{T}$, forcing the world model to guess among multiple potential future states.
  In \textbf{(b)}, starting from the top left, exploratory actions $a_{1:T_c}^c$ are first executed in the test scene $\mathcal{T}$, yielding observations ${o}_{1:T_c}^c$. The context encoder $g_\phi$ then extracts physics information from this obs-action sequence and compresses it into a physics descriptor token \zp. Conditioned on $\textbf{z}_{\text{phy}}$, the world model is able to generate future predictions $\hat{o}_{t+1}$ consistent with the test scene's dynamics.
}
\vspace{-15pt}
\label{fig:2-method}
\end{figure}

\subsection{Physics Calibration via Physics Descriptor Token}
\label{sec:method:descriptor_token}
\paragraph{Physics Ambiguity in Visual World Models.}
The forward transition function $\mathcal{T}$ of an environment depends not only on the current states and actions, but also on a set of underlying physics properties $\xi \in \mathcal{X}$ (e.g., mass, surface friction, or joint damping). 
Because many of these physics properties rarely correlate with surface appearance,
a static visual frame $o_t$ or a brief observation window can hardly identify $\xi$ and hence leaves the true transition dynamics underdetermined.

Consequently, action-conditioned future prediction becomes an ill-posed, one-to-many problem.
Without sufficient interaction data from the test scene to calibrate the future predictor $f_\theta$, its prediction may collapse to the expectation over the training distribution with compounding dynamic drift when the test environment deviates from the nominal mean. 
While one could explicitly estimate and feed the test-scene physics parameters to $f_\theta$, this usually relies on ad-hoc analytical modeling (as one must know in advance which physics properties determine the dynamics) and also introduces cascading estimation errors.
An alternative strategy is to scale up the history window $K$ of world models, so that the future predictor $f_\theta$ can directly extract the physical transition dynamics from previous interaction. 
However, self-attention complexity in transformers scales quadratically with sequence length, making long-history conditioning computationally prohibitive. Furthermore, continuously carrying long raw frame histories (e.g., thousands of tokens) in every inference step introduces substantial redundancy, as the underlying physics configuration can usually be represented far more compactly.

\paragraph{Physics Descriptor Token.}
To address these challenges, our method calibrates world models through a single compact representation, termed \textit{physics descriptor token} \zp, which is extracted from an exploratory interaction sequence collected in the test scene. 
Formally, given an interaction clip $\mathcal{C} = \{(o_i^c, a_i^c)\}_{i=1}^{T_c}$ of length $T_c$, the context encoder \ctxenc maps it to a descriptor token 
$\zp = \ctxenc (\mathcal{C}) \in \mathbb{R}^{d_\text{phy}}$. 
Internally, a lightweight visual encoder $E_c^v$ extracts per-frame visual features and an action encoder $E_c^a$ maps the control actions into action embeddings. 
At each timestep $i$, the two are concatenated along the channel dimension into a joint transition token $[E_c^v(o_i^c) \mid E_c^a(a_i^c)] \in \mathbb{R}^{d_v + d_a'}$. 
A transformer $E_c^\text{fuse}$ then fuses these transition tokens across time and pools them via a learnable query token $\mathbf{z}_\text{cls}$:
\begin{equation}
\zp
= \ctxenc(\mathcal{C})
\coloneqq 
E_c^\text{fuse}
\Bigl(
  \mathbf{z}_{\text{cls}},\;
  \bigl\{ \bigl[\, E_c^v(o_i^c) \;\big|\; E_c^a(a_i^c) \,\bigr] \bigr\}_{i=1}^{T_c}
\Bigr)_0,
\end{equation}
where $[\,\cdot \mid \cdot\,]$ denotes channel-wise concatenation and $(\cdot)_0$ denotes the output at the query-token position. We use $\phi$ to collectively denote the parameters of \ctxenc, including $E_c^\text{fuse}$, $E_c^v$, $E_c^a$, and $\mathbf{z}_\text{cls}$.
The world model's latent dynamics predictor $f_\theta$ is subsequently conditioned on
\zp alongside the default observation-action history window $K$, predicting future latents $\hat{z}_{t+1}$ as:
\vspace{-2pt}
\begin{equation}
    \hat{z}_{t+1} = f_\theta\Bigl( \bigl\{(z_i, e_i^a)\bigr\}_{i=t-K+1}^{t},\; \zp \Bigr).
\vspace{-5pt}
\end{equation}

\textbf{Efficiency and Reusability.} The design of \textit{physics descriptor token} provides two practical advantages. (1) \emph{minimal rollout overhead}: adding a single additional token \zp into the predictor $f_\theta$ introduces only a constant and almost negligible computational cost, circumventing the quadratic attention overhead compared to using longer history or simply prepending raw context frames to $f_\theta$ (see \textit{+Long-History} and \textit{+Context-Prefix} in Section \ref{sec:experiments} and \autoref{fig:7-inference-time-main} for comparison);
(2) \emph{one-pass extraction and in-context reuse}: \zp is computed via a single forward pass of a lightweight context encoder and can then be cached.
If the dynamics of the test scene remain unchanged, the same descriptor can be reused across arbitrarily many future rollouts or planning tasks with no additional encoding cost.

\subsection{Training \method without Physics Labels}\label{sec:method:training_without_physics_labels}
To enable physics-calibrated predictions, existing models typically rely on pre-specified analytical equations~\citep{li2025pin,wang2026contactgaussian} or train explicit system identification modules using privileged physical parameters~\citep{yu2017preparing,kumar2021rma}. 
However, constructing analytical physics formulations often demands substantial human effort and
is often intractable for unstructured scenes with coupled dynamics.
Meanwhile, obtaining the ground-truth physics parameters requires instrumented environments and often introduces measurement errors, which would then propagate to model training. As a result, these approaches are typically confined to controlled settings and struggle to scale to novel, uncalibrated in-the-wild scenes governed by unknown physical properties.

In contrast, \method requires neither prior physics equations nor the error-prone explicit parameter estimation during both training and evaluation. 
While deriving analytical formulations or measuring exact physical parameters of in-the-wild scenes is intractable, 
constructing data groups where each group consists of interaction data collected within the same physical environment is simple and requires minimal additional effort during data collection.
Such data grouping alone is sufficient to train \method and learn an effective \textit{physics descriptor token}.

Specifically, the training dataset is categorized into $M$ groups $\mathcal{D}_{\text{train}} = \{\mathcal{G}_j\}_{j=1}^M$, where each group $\mathcal{G}_j$ comprises episodes that share the same underlying physics configuration $\xi_j$.
In training, for any transition $(o_t, a_t, o_{t+1}) \sim \mathcal{G}_j$, we independently sample another clip $\mathcal{C}_{\text{tr}}^j = \{(o_k^c, a_k^c)\}_{k=1}^{T_c} \sim \mathcal{G}_j$ from the same group as the interaction prompt.
Because both sequences share the identical underlying transition dynamics $\mathcal{T}_{\xi_j}$, the context encoder \ctxenc can learn to distill the scene's physical characteristics into the compact descriptor token \zp to synthesize the correct prediction. 
We jointly optimize the context encoder $g_\phi$ and dynamics predictor $f_\theta$ end-to-end using the latent prediction objective:
\begin{equation}
\label{eq:wm_training_loss}
\begin{aligned}
\mathcal{L}(\theta, \phi)
&= \mathbb{E}_{\mathcal{G}_j \sim \mathcal{D}_{\text{train}},\; (o_t, a_t, o_{t+1}) \sim \mathcal{G}_j,\; \mathcal{C}_{\text{tr}}^{j} \sim \mathcal{G}_j}
   \Bigl[\, \bigl\| f_\theta\bigl(z_t, e_t^a, \zp\bigr) - z_{t+1} \bigr\|_2^2 \,\Bigr], \\
&\quad \text{where}~~
   z_t = E_\text{v}(o_t),~~
   e_t^a = E_\text{a}(a_t),~~
   z_{t+1} = E_\text{v}(o_{t+1}),~~
   \zp = \ctxenc\bigl(\mathcal{C}_{\text{tr}}^{j};\, \phi\bigr).
\end{aligned}
\end{equation}
For simplicity, we assume the input window size $K$ of $f_\theta$ to be 1 in \autoref{eq:wm_training_loss}.
This formulation enables \method to extract instance-specific physical dynamics directly from the interaction clip $\mathcal{C}_{\text{tr}}$ into the compact physics descriptor token \zp, circumventing the need for ground-truth physics labels as supervision.
As in \autoref{fig:8-ctx-token}, this self-supervised predictive objective can autonomously learn a structured latent space correlated to the underlying physics variations of the scenes.

\subsection{Inference with \method}\label{sec:method:inference}

\paragraph{Collecting the Interaction Prompt.}
At test time, \method enables zero-shot adaptation to environments with novel physics configurations without fine-tuning. 
To calibrate the world model to an unseen test scene, the agent first executes a sequence of exploratory actions to collect a context prompt $\mathcal{C}_{\text{eval}} = \{(o_k^e, a_k^e)\}_{k=1}^{T_c}$ of length $T_c$, which consists of the applied actions and the corresponding observations. The actions $\{a_k^e\}_{k=1}^{T_c}$ can be either randomly sampled or generated by a task-agnostic playful policy, as this interaction prompt is only for revealing the physical dynamics of the scene.
However, these actions should produce physical contact with the environment since interactions lacking contact or motion cannot reveal the underlying dynamics.

\paragraph{Physics-grounded Rollout.}
Given $\mathcal{C}_\text{eval}$ collected in the test scene, the physics descriptor token is computed once by the context encoder $g_\phi$ and can be held fixed if the test scene dynamics remain unchanged. Calibrated by descriptor $\zp^\text{ev}$, the predictor $f_\theta$ can then predict the future latents as:
\begin{equation}
\label{eq:inference}
    \hat{z}_{t+1} = f_\theta\Bigl( \bigl\{(z_i, e_i^a)\bigr\}_{i=t-K+1}^{t},\; \zp^\text{ev} \Bigr),
    \quad
    \zp^\text{ev} = \ctxenc(\mathcal{C}_\text{eval}),
\end{equation}
where predicted latents $\hat{z}_i$ can be fed back to $f_\theta$ as observation to enable long-horizon open-loop rollout (see Algorithm \ref{alg:icp_inference} for pseudocode). If the intrinsic physics of the scene does not change, \zp can be shared across every rollout step and across every candidate trajectory to be evaluated during visual planning.
Therefore, the prompt only needs to be encoded once per scene, regardless of the planning horizon or the number of sampled action sequences.
Notably, no physical parameters, analytical models, or task-specific fine-tuning are required at inference: adapting \method to a new scene only requires collecting a period of in-context interaction.

%% file: sections/experiments.tex
\vspacet{-3pt}
\section{Experiments} \label{sec:experiments}
\vspacet{-3pt}
\begin{figure}[t]
  \centering
  \includegraphics[width=1.0\linewidth]{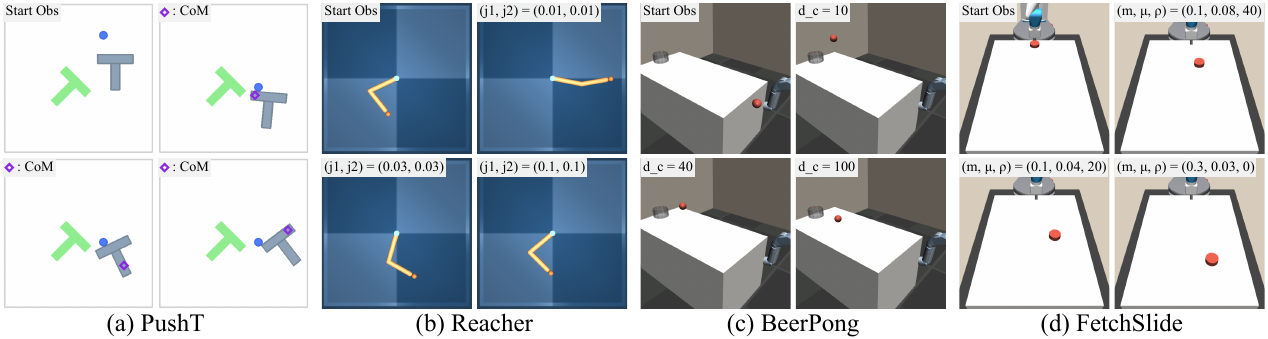}
  \vspace{-17pt}
  \caption{\textbf{Environment Demonstrations.} 
  We evaluate our method on four simulation environments featuring diverse variations of physics properties. For each environment, we show four images: the top-left image is a random initial observation, while the remaining three images demonstrate how executing the \textit{same} action sequence yields different future states due to different physics parameters of the scenes.
  Specifically, in (a) Push-T, the center of mass of the T-block is varied; in (b) Reacher, the joint damping of the two-link arm is varied; in (c) BeerPong, the energy dissipation upon table contact is varied; in (d) FetchSlide, \textit{three} physics properties vary concurrently: puck mass, surface friction, and aerodynamic drag.
  }
  \label{fig:4-env-demo}
  \vspace{-9pt}
\end{figure}

Our experiments evaluate five questions:
\textbf{(1)} \text{Physics Calibration:} Does conditioning on in-context interaction clips enable effective physics calibration and yield more accurate future predictions?
\textbf{(2)} \text{Descriptor Expressiveness and Efficiency:} Can \method learn compact, expressive \textit{physics descriptor tokens} without analytical physics priors and enable high compute efficiency?
\textbf{(3)} \text{Coupled Physics Dynamics:} Can \method scale to complex scenes whose dynamics are jointly governed by multiple physics properties?
\textbf{(4)} \text{Backbone Generalization:} Can \method function as an architecture-agnostic, plug-and-play module across diverse world-model backbones?
\textbf{(5)} \text{Out-of-Distribution Extrapolation:} Can \method extrapolate to unseen environments whose physics parameters lie outside the training distribution?

\textbf{Baselines.} Our in-context prompting framework can be integrated into various world model backbones. We use DINO-WM~\citep{zhou2025dino} and LeWM~\citep{maes2026leworldmodel} as our base world models due to their rollout efficiency and strong trajectory planning performance, alongside Temporal Straightening (TS-WM)~\citep{wang2026temporal} as a competitive latent world model baseline.

We train several variants of models for systematic study: 
\textbf{(1)} \textit{+Long History}, which takes in a long history window $K$ to investigate if simply extending  history memory in world models can enable physics-calibrated predictions;
\textbf{(2)} \textit{+Context Prefix}, which directly prepends the raw context interaction prompt to the input of the world model instead of using the proposed physics descriptor token;
\textbf{(3)} \text{\textit{+Estimated Physics}$^\dagger$}, 
which trains an encoder to predict physics parameters from a context clip and feeds the estimated physics values to the world model (hence this baseline requires additional physics labels for model training).
For further analysis, we additionally evaluate \textbf{(4)} \textit{+Privileged Physics}$^\ddagger$ in Section \ref{sec:experiment:analysis}, which takes the ground-truth test-scene physics parameters as input, serving as an oracle ceiling rather than a directly comparable baseline. 
Finally, we compare LeWM against LeWM+\shortmethod to verify that our method generalizes across different world model backbones. See Appendix~\ref{sec:appendix_baselines} for detailed description of each baseline.

\begin{figure}[t]
  \centering
  \includegraphics[width=1.0\linewidth]{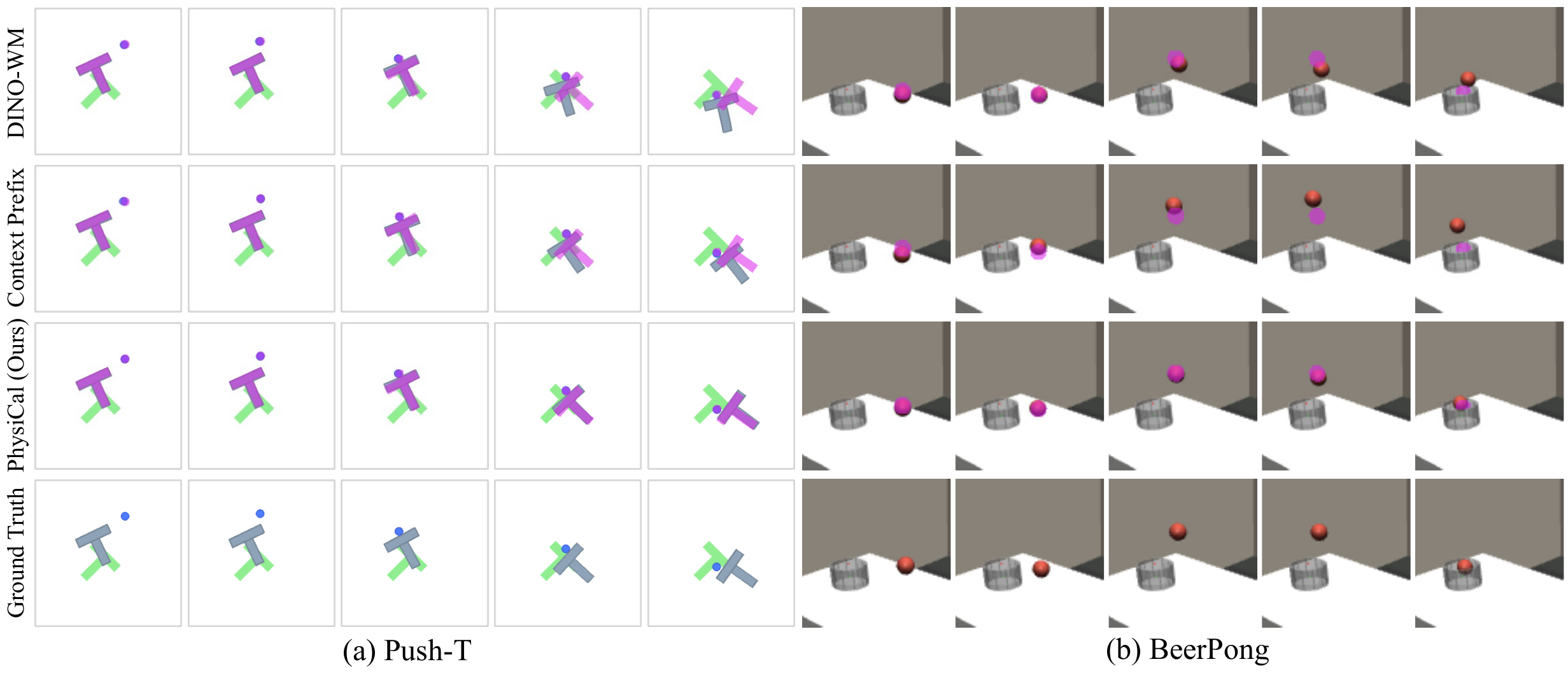}
  \vspace{-20pt}
  \caption{\textbf{Qualitative Open-Loop Rollout Results in Push-T and BeerPong.} We show a subset of the full rollouts due to limited space and crop the synthesized images in BeerPong for clearer view. 
  We use \pinkbox~masks to indicate the ground-truth location of the target objects. 
Note that the physics configuration of these test scenes is unknown to the models and \shortmethod calibrates the predictions solely based on a given in-context interaction clip (\autoref{fig:context-main-pusht} and \autoref{fig:context-main-beerpong}), yielding results that closely match the ground truth.
  }
  \label{fig:5-main-pusht-fetchslide}
  \vspacet{-9pt}
\end{figure}

\begin{table}[t]
    \centering
    \small
    \setlength{\tabcolsep}{4pt}
    \adjustbox{max width=\textwidth}{
    \begin{tabular}{l c c c c c c c c c c}
        & \multicolumn{5}{c}{\textbf{Push-T}} & \multicolumn{5}{c}{\textbf{Reacher}} \\
        \cmidrule(lr){2-6} \cmidrule(lr){7-11}
        \textbf{Method} & PSNR$\uparrow$ & SSIM$\uparrow$ & LPIPS$\downarrow$ & IoU$\uparrow$ & SR(\%)$\uparrow$ & PSNR$\uparrow$ & SSIM$\uparrow$ & LPIPS$\downarrow$ & IoU$\uparrow$ & SR(\%)$\uparrow$ \\
        \midrule
        TS-WM & 21.38 & 0.954 & 0.0405 & 53.4 & 53.3 & 19.51 & 0.967 & 0.0469 & 30.9 & 41.3 \\
        DINO-WM & 21.87 & 0.955 & 0.0402 & 51.5 & 53.3 & 19.43 & 0.968 & 0.0407 & 32.6 & 52.0 \\
        ~ + Long History & 22.24 & 0.958 & 0.0342 & 55.9 & 54.7 & 20.44 & 0.972 & 0.0302 & 41.5 & 56.0 \\
        ~ + Context Prefix & 22.98 & \underline{0.961} & \underline{0.0295} & \underline{61.4} & \underline{58.7} & 20.13 & 0.970 & 0.0332 & 38.8 & \underline{61.3} \\
        ~ + Estimated Physics$^\dagger$ & \underline{23.11} & \underline{0.961} & 0.0319 & 59.0 & 57.3 & \textbf{21.82} & \textbf{0.976} & \textbf{0.0221} & \textbf{51.5} & \textbf{69.3} \\
        \rowcolor{green!17}
        ~ + \shortmethod (Ours) & \textbf{23.83} & \textbf{0.964} & \textbf{0.0267} & \textbf{65.1} & \textbf{62.7} & \underline{21.58} & \underline{0.975} & \underline{0.0222} & \underline{51.0} & \underline{61.3} \\
        \midrule
        LeWM & \underline{23.72} & \underline{0.957} & \underline{0.0450} & \underline{64.3} & \underline{48.0} & \underline{20.45} & \textbf{0.964} & \underline{0.0404} & \underline{40.6} & \underline{60.0} \\
        \rowcolor{green!17}
        ~ + \shortmethod (Ours) & \textbf{25.50} & \textbf{0.964} & \textbf{0.0358} & \textbf{73.3} & \textbf{61.3} & \textbf{21.08} & \textbf{0.964} & \textbf{0.0377} & \textbf{45.7} & \textbf{64.0} \\
        \bottomrule
    \end{tabular}
    }
    \vspace{-7pt}
    \caption{
    \textbf{Push-T with Varying Center of Mass of the T-block:} We evaluate on 15 unseen center-of-mass (CoM) configurations and report the quantitative visual metrics of open-loop rollout and planning success rates (SR). 
    IoU is computed on the combined mask of the agent and the T-block.
    Different CoMs will significantly affect the motion pattern given the same contact, hence lack of physics calibration will result in low IoU.
    \textbf{Reacher with Different Joint Damping:} The robot arm contains 2 joints and we evaluate on 15 unseen joint damping combinations.
    Different damping values alter resistance to motion, resulting in different displacement amplitude of the arm under identical torques (see \autoref{fig:4-env-demo}). 
    }
    \label{tab:main_pusht_reacher_results}
    \vspacet{-13pt}
\end{table}

\textbf{Environments and Evaluation.}
We illustrate all our environments in \autoref{fig:4-env-demo} and show how different physics configurations will affect the future world states. We train the models on $M=135$ different configurations and report the results on 15 unseen configurations, with 5 validation episodes per configuration, giving 75 episodes in total.
We report four visual metrics (PSNR, SSIM, LPIPS, IoU) in open-loop rollout, which starts from the initial frame and autoregressively predicts all frames of an episode by recursively feeding the model's own predictions back as input.
We also evaluate on 75 planning tasks per environment and report the success rate (SR). Each task is defined by a start image and a goal image. We execute the planned actions in the environments. The success criterion is whether the target object reaches the goal within a threshold. See Appendix~\ref{sec:appendix_environments} for details.

\vspacet{-3pt}
\subsection{Environments with Single Physics Variation}
\paragraph{Push-T under Varying Center of Mass.} The Push-T environment consists of a circular pusher (the agent) and a T-shaped block (see \autoref{fig:4-env-demo}). The world model predicts the future latents of the scene alongside the pusher's proprioception conditioned on the pusher's actions. In planning, each goal is defined by an image and a target agent state.
The block's center of mass (CoM) is randomly sampled. Different CoMs will not change the block's appearance but will alter how the block rotates and slides when pushed, hence resulting in significantly different states even under identical push actions.

\begin{figure}[t]
    \vspace{-6pt}
    \centering
    \begin{minipage}{0.49\linewidth}
        \centering
        \includegraphics[width=\linewidth]{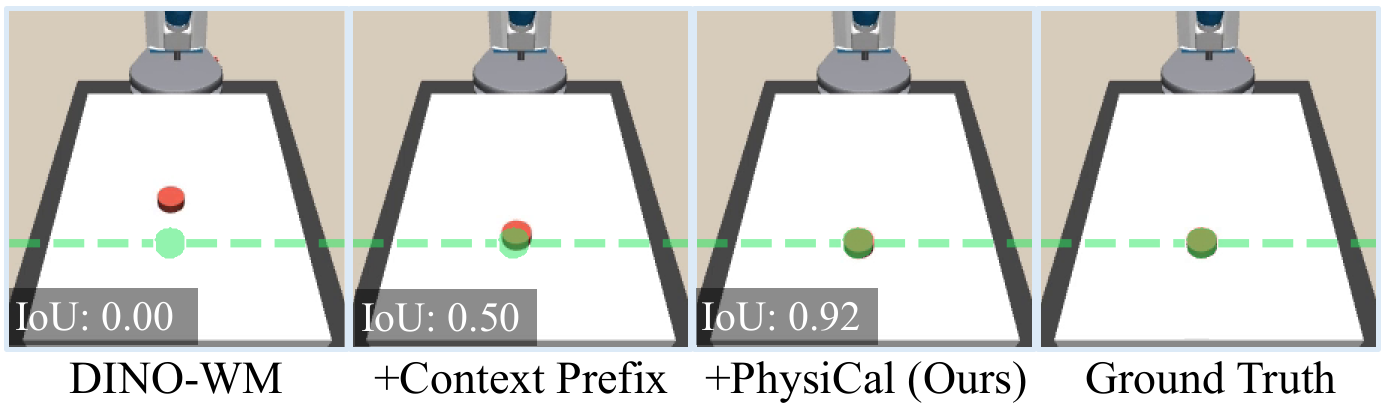}
        \vspace{-20pt}
        \caption{\small \textbf{Qualitative Comparison in FetchSlide.} 
        Visual comparison of open-loop rollout predictions under identical striking actions in an unseen test scene. We only demonstrate the final frame due to space limit. The green mask denotes the ground-truth puck position.
        }
        \label{fig:6-fetchslide-main}
    \end{minipage}
     \hfill
    \begin{minipage}{0.5\linewidth}
        \centering
        \includegraphics[width=\linewidth]{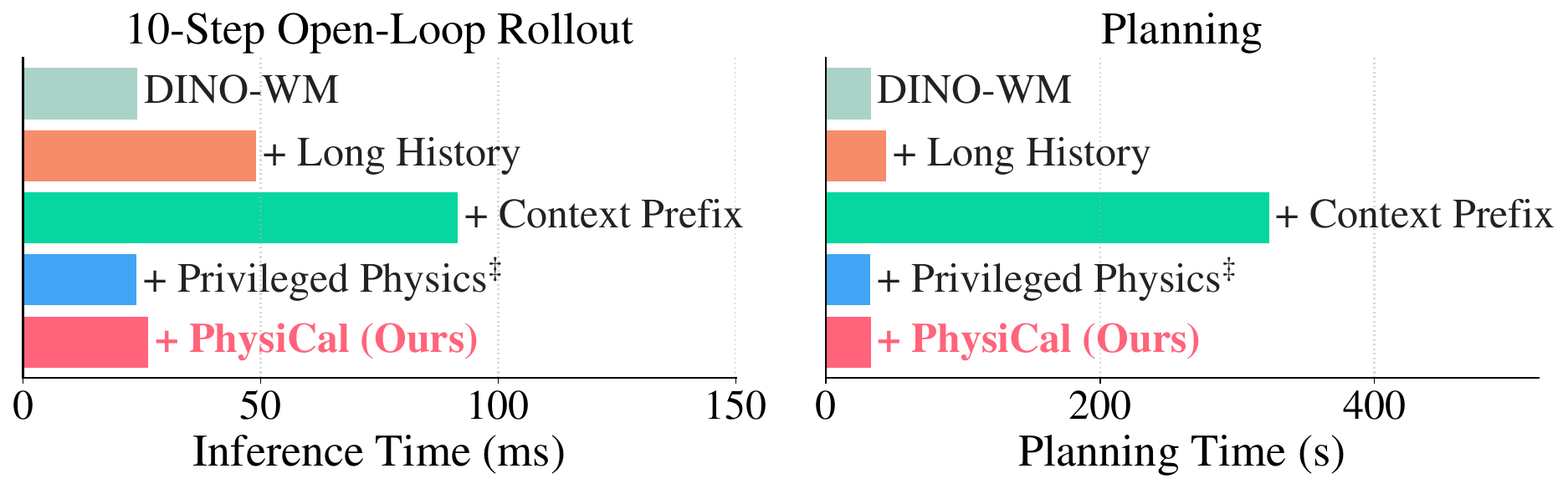}
        \vspace{-18pt}
        \caption{\small 
        \textbf{Inference Speed Comparison of Open-Loop Rollout and Planning in Push-T.} 
        For planning, we use the same  Cross-Entropy Method (CEM) setup as DINO-WM. See Appendix \ref{sec:appendix_training_time} and \ref{sec:appendix_inference_time} for additional results.
        }
        \label{fig:7-inference-time-main}
    \end{minipage}
    \vspace{-16pt}
\end{figure}

As in \autoref{tab:main_pusht_reacher_results}, methods without in-context prompts as input (e.g., DINO-WM and TS-WM) generally underperform, showing that in-context interaction prompts can effectively facilitate physics calibration.
We observe that compared to other long-context variants (e.g., +Long History and +Context Prefix, see Appendix~\ref{sec:appendix_baselines} for details), +\shortmethod achieves significantly better performance, especially in challenging open-loop rollout, showing that a single descriptor token can stably ground the physics over long-horizon rollout.
Our method also outperforms +Estimated Physics$^\dagger$ in all open-loop rollout visual metrics without using any physics labels during training. Moreover, compared to +Context Prefix, our compact descriptor token design avoids the quadratic self-attention cost caused by the long context sequence, yielding markedly speedup in both training and inference (\autoref{fig:7-inference-time-main} and \autoref{tab:training_time_pusht_reacher}).

\paragraph{Reacher Control under Varying Joint Damping.}
This environment consists of a two-link robot arm actuated by continuous joint torques. 
The damping coefficients of the two joints can affect how much the arm moves under the same applied torques. 
Each planning task optimizes a torque trajectory to steer the arm toward a target state specified by a goal image. The in-context prompts are randomly sampled actions. 
We present the quantitative results in \autoref{tab:main_pusht_reacher_results}. 
Our in-context conditioning consistently improves both DINO-WM and LeWM, outperforming all unprivileged baselines and trailing only \textit{+Estimated Physics}, which requires ground-truth joint damping parameters for model training.
While +\shortmethod achieves a similar planning success rate as \textit{+Context Prefix}, our method  surpasses it across all visual prediction metrics by a margin and is more compute efficient.
We specify implementation details of the Reacher environment in Appendix~\ref{sec:appendix_reacher}.

\begin{table}[t]
    \centering
    \small
    \setlength{\tabcolsep}{3pt}
    \adjustbox{max width=\textwidth}{
    \begin{tabular}{l c c c c c c c c c c}
        & \multicolumn{5}{c}{\textbf{BeerPong}} & \multicolumn{5}{c}{\textbf{FetchSlide}} \\
        \cmidrule(lr){2-6} \cmidrule(lr){7-11}
        \textbf{Method} & PSNR$\uparrow$ & SSIM$\uparrow$ & LPIPS$\downarrow$ & IoU$\uparrow$ & SR (\%)$\uparrow$ & PSNR$\uparrow$ & SSIM$\uparrow$ & LPIPS$\downarrow$ & IoU$\uparrow$ & SR (\%)$\uparrow$ \\
        \midrule
        TS-WM & 26.97 & 0.992 & 0.0129 & 24.8 & 60.0 & 24.07 & 0.984 & 0.0131 & 49.9 & 26.7 \\
        DINO-WM & 27.35 & 0.993 & 0.0125 & 27.7 & 44.0 & 25.67 & 0.989 & 0.0077 & 58.6 & 26.7 \\
        ~ + Long History & 27.11 & 0.993 & 0.0138 & 24.8 & 50.7 & 25.17 & 0.989 & 0.0090 & 51.0 & 24.0 \\
        ~ + Context Prefix & 29.10 & \underline{0.994} & 0.0089 & 40.8 & \textbf{82.7} & \underline{27.14} & \textbf{0.991} & \underline{0.0048} & \underline{71.5} & 54.7 \\
        ~ + {Estimated Physics}$^\dagger$ & \textbf{29.58} & \textbf{0.995} & \underline{0.0074} & \underline{44.0} & 77.3 & 26.93 & \underline{0.990} & 0.0052 & 69.6 & \textbf{69.3} \\
        \rowcolor{green!17}
        ~ + \shortmethod (Ours) & \underline{29.49} & \textbf{0.995} & \textbf{0.0067} & \textbf{46.3} & \underline{80.0} & \textbf{27.87} & \textbf{0.991} & \textbf{0.0047} & \textbf{73.8} & \underline{56.0} \\
        \midrule
        LeWM$^*$ & \underline{25.05} & \underline{0.983} & \underline{0.0230} & \underline{27.2} & \underline{60.0} & \underline{25.15} & \underline{0.981} & \underline{0.0336} & \underline{51.1} & \underline{17.3} \\
        \rowcolor{green!17}
        ~ + \shortmethod (Ours) & \textbf{27.58} & \textbf{0.986} & \textbf{0.0170} & \textbf{47.7} & \textbf{84.0} & \textbf{26.43} & \textbf{0.982} & \textbf{0.0321} & \textbf{61.8} & \textbf{28.0} \\
        \bottomrule
    \end{tabular}
    }
    \vspace{-7pt}
    \caption{
   \textbf{BeerPong with Different Ball-contact Damping.}
  In BeerPong, we vary table contact damping, which defines the ball's rebound height and trajectory.
  For planning, the agent needs to throw the ball such that it bounces off the table and lands inside the target cup. IoU is computed on the ball foreground mask.
\textbf{FetchSlide with Multiple Physics Variations.}
    This environment simultaneously varies three physical properties: puck mass, table surface friction, and air drag, which jointly determine how the puck slides. In planning, the agent is required to strike the puck to a given location.
    Estimated Physics$^\dagger$ requires environment physics configurations for model training, while ours does not. LeWM$^*$: we use TC-LeWM in BeerPong to enhance performance (see Appendix \ref{sec:appendix_baselines} for details).
   }
    \label{tab:main_beerpong_fetchslide_results}
\end{table}

\paragraph{BeerPong under Varying Impact Energy Dissipation.}
In BeerPong, a robot arm throws a ball toward a target cup on a tabletop. 
The ball needs to bounce off the table at least once to enter the cup due to the long distance.
We vary impact energy dissipation by changing the table's contact damping (e.g., higher damping causes greater energy loss, resulting in lower bounce height).
The action input to the world models is the end-effector release velocity. The cup location is randomly initialized in each planning task. The planner then optimizes the release velocity so that the ball can land right inside the cup after bounce. 
The in-context prompts are collected via randomly sampled actions.
Please see Appendix~\ref{sec:appendix_beerpong} for additional details.

We present quantitative results in \autoref{tab:main_beerpong_fetchslide_results}. Across both world-model backbones, \text{+\shortmethod} consistently improves visual prediction accuracy over the uncalibrated baselines. While our method surpasses \textit{+Context Prefix} across all visual metrics, it achieves a slightly lower planning success rate.
This is potentially because the physics variation in BeerPong manifests only when the ball contacts the table, leading to temporally sparse physical cues. Under such observation sparsity, self-attention over the full raw context frames may provide an advantage in identifying collision dynamics.

\begin{figure}
\begin{minipage}[t]{0.49\textwidth}
    \vspace{0pt}
    \centering
    \footnotesize
    \setlength{\tabcolsep}{2pt}
    \adjustbox{max width=\linewidth}{%
    \begin{tabular}{l*{6}{c}}
    & \multicolumn{3}{c}{\textbf{Push-T}} & \multicolumn{3}{c}{\textbf{Reacher}} \\
    \cmidrule(lr){2-4}\cmidrule(lr){5-7}
    Method & PSNR$\uparrow$ & IoU$\uparrow$ & SR$\uparrow$ & PSNR$\uparrow$ & IoU$\uparrow$ & SR$\uparrow$ \\
    \midrule
    DINO-WM & 21.87 & 51.5 & 53.3 & 19.43 & 32.6 & 52.0 \\
    + Privileged Physics$^\ddagger$ & \underline{23.36} & \underline{60.9} & \underline{60.0} & \textbf{22.92} & \textbf{59.3} & \textbf{86.7} \\
    \rowcolor{green!17}
    + \shortmethod (Ours) & \textbf{23.83} & \textbf{65.1} & \textbf{62.7} & \underline{21.58} & \underline{51.0} & \underline{61.3} \\
    \midrule
    & \multicolumn{3}{c}{\textbf{BeerPong}} & \multicolumn{3}{c}{\textbf{FetchSlide}} \\
    \cmidrule(lr){2-4}\cmidrule(lr){5-7}
    Method & PSNR$\uparrow$ & IoU$\uparrow$ & SR$\uparrow$ & PSNR$\uparrow$ & IoU$\uparrow$ & SR$\uparrow$ \\
    \midrule
    DINO-WM & 27.35 & 27.7 & 44.0 & 25.67 & 58.6 & 26.7 \\
    + Privileged Physics$^\ddagger$ & \textbf{30.27} & \textbf{50.3} & \textbf{86.7} & \underline{27.68} & \textbf{75.8} & \textbf{66.7} \\
    \rowcolor{green!17}
    + \shortmethod (Ours) & \underline{29.49} & \underline{46.3} & \underline{80.0} & \textbf{27.87} & \underline{73.8} & \underline{56.0} \\
    \bottomrule
\end{tabular}}
    \vspace{-9pt}
    \captionof{table}{\textbf{Quantitative Comparison of +\shortmethod and Privileged Physics$^\ddagger$.} +\shortmethod calibrates physics only through an interaction clip collected in the test scene, while +Privileged Physics$^\ddagger$ directly uses the ground-truth physics parameters of the test scene as input to calibrate the world models.
    }
    \label{tab:main-ours-privileged}
\end{minipage}%
\hfill
\begin{minipage}[t]{0.49\textwidth}
    \vspace{0pt}
    \centering
    \includegraphics[width=\linewidth]{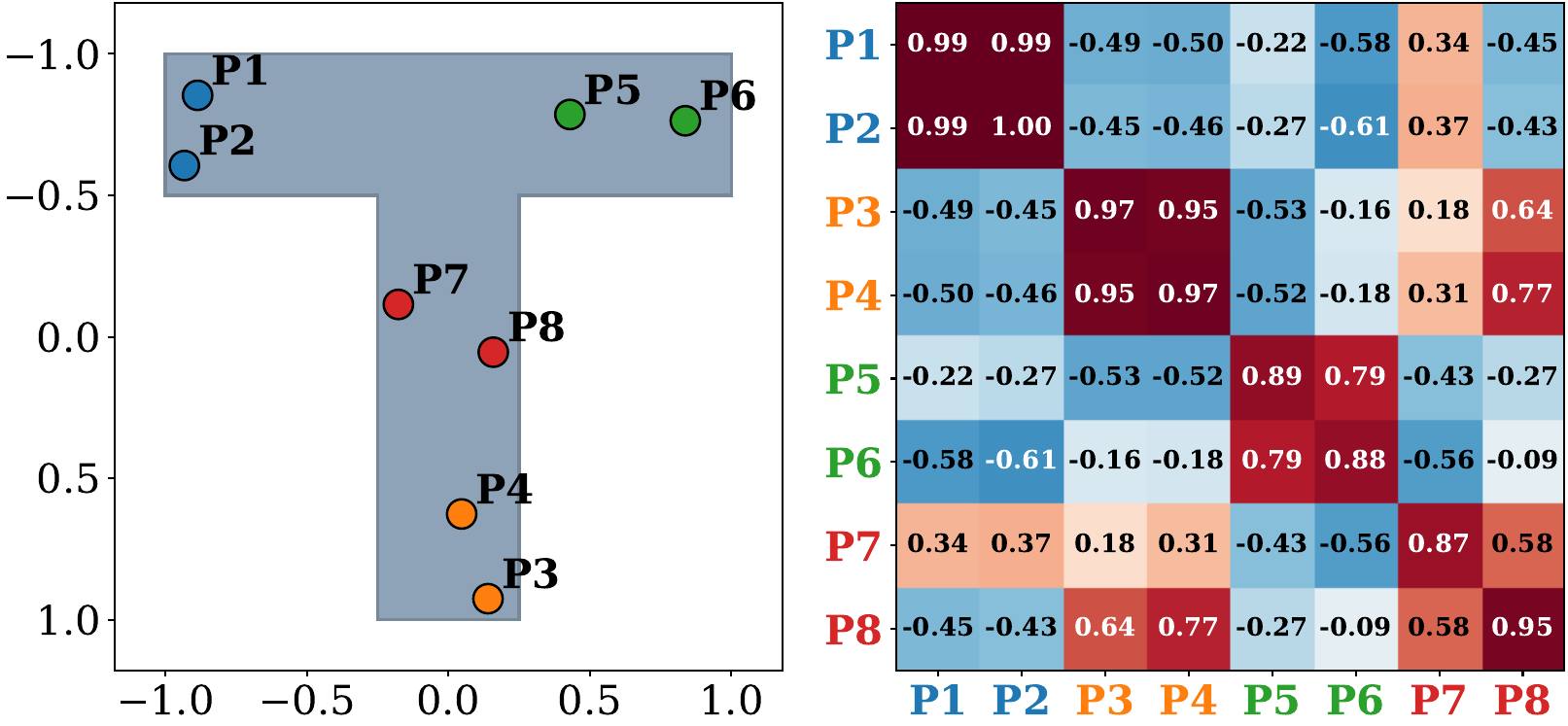}
    \vspace{-17pt}
    \captionof{figure}{\textbf{Cosine Similarity of Physics Descriptor Tokens of Different CoMs in Push-T.} We compute descriptor token \zp of 8 unseen centers of mass and visualize their pairwise cosine similarity. Trained without any physics labels, our method can still learn a structured, physics-correlated latent space, e.g., adjacent centers of mass yield high similarity while remote pairs yield negative values.}
    \label{fig:8-ctx-token}
\end{minipage}
\vspace{-14pt}
\end{figure}

\vspacet{-2pt}
\subsection{Environments with Multiple Physics Variations}
\textbf{FetchSlide under Multiple Physics Variations.}
We next investigate whether \shortmethod can calibrate world models in environments where transition dynamics are simultaneously governed by \textit{multiple} physics properties. 
To this end, in FetchSlide, we concurrently vary three physics parameters: puck mass, table surface friction, and aerodynamic drag. In this environment, a robot arm strikes a flat puck across a tabletop, where these three properties jointly determine the puck's post-strike deceleration and sliding distance.
The action space is end-effector delta position, which subsequently defines the striking velocity and angle. In planning, we optimize the strike actions so that the puck can stop at the target position specified by a goal image. The context sequence is collected via random actions.

Quantitative results are presented in \autoref{tab:main_beerpong_fetchslide_results}. Despite the added complexity of coupled physical factors, +\shortmethod effectively calibrates predictions without requiring analytical physics equations, 
\begin{wrapfigure}{r}{0.5\textwidth}
    \centering
    \includegraphics[width=\linewidth]{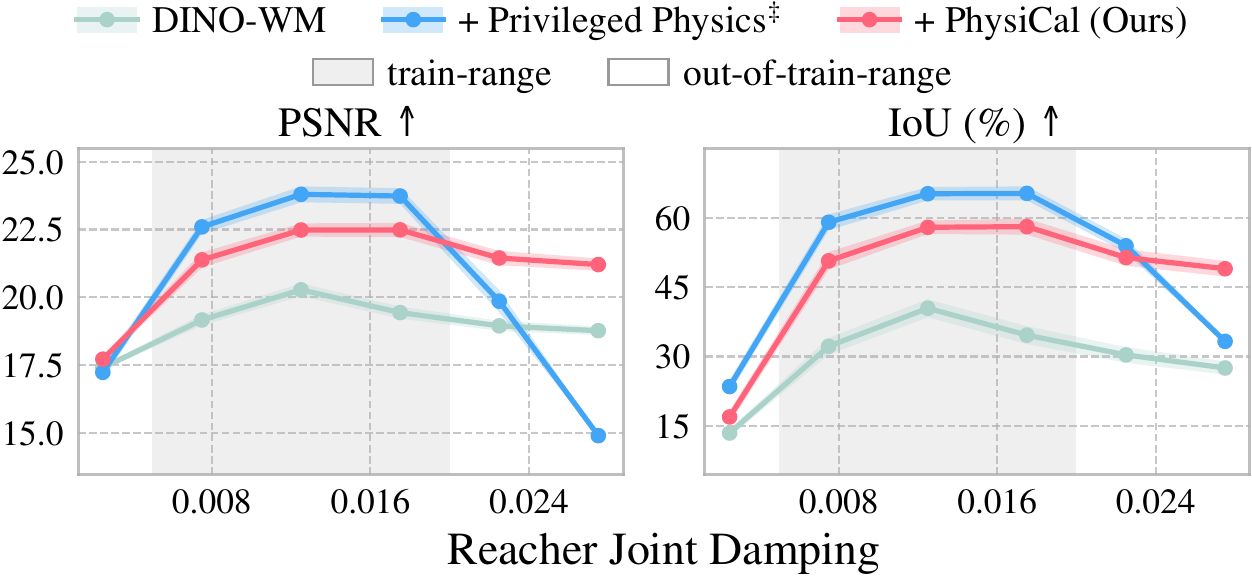} 
    \vspace{-18pt}
    \caption{\textbf{Extrapolation to Out-of-Distribution Physics Parameters in Reacher.} We report results of open-loop rollouts. For each circle, we sample 15 damping values within a bin centered at its $x$-coordinate evaluated on 5 episodes per value, 75 total.}
    \label{fig:9-extrapolate}
    \vspace{-16pt}
\end{wrapfigure}
yielding significant gains across both backbones.
Our method is only surpassed by \textit{+Estimated Physics}$^\dagger$, which is trained with the ground-truth values of all three physical properties. 
However, oftentimes finding what physics parameters govern the scene and precisely measuring them are impractical, whereas +\shortmethod is trained directly via the future latent prediction objective without any physics labels. 
We also provide qualitative results of the final prediction of an open-loop rollout in \autoref{fig:6-fetchslide-main}, where uncalibrated DINO-WM deviates from the target position, while +\shortmethod generates predictions that faithfully capture the ground-truth puck state.

\vspacet{-2pt}
\subsection{Analysis}\label{sec:experiment:analysis}
\vspacet{-2pt}

\textbf{Physics Calibration \textit{vs.} Privileged Physics$^\ddagger$.} 
We additionally compare our method against an oracle baseline, \textit{+Privileged Physics}$^\ddagger$, serving as an upper-bound reference. Specifically, this baseline directly receives ground-truth physics parameters as supplementary inputs during both training and inference (Privileged Physics$^\ddagger$ and Estimated Physics$^\dagger$ share the same future predictor $f_\theta$, but at inference, Estimated Physics$^\dagger$ uses physics parameters estimated by a neural network). 
In contrast, +\shortmethod accesses only an unannotated interaction clip from the test scene for calibration, and nevertheless, as shown in Table \ref{tab:main-ours-privileged}, our method achieves competitive performance across most metrics and even surpasses this privileged baseline in Push-T.

\textbf{Generalization to Out-of-Distribution Physics Parameters.} We investigate whether \method can extrapolate to environments where physics parameters lie outside the training distribution (\autoref{fig:9-extrapolate}). 
Specifically, in Reacher, we sweep the first joint's damping coefficient beyond its training range while holding the second joint's damping fixed. 
While \textit{+Privileged Physics} achieves higher accuracy in-distribution, its performance deteriorates sharply when the damping falls below or exceeds the training interval. 
In particular, its rollout PSNR significantly drops, producing blurry predictions (\autoref{fig:reacher-extrapolation-qualitative}).
In contrast, our method demonstrates robust extrapolation over unseen regimes, probably because our model learns latent patterns to infer dynamics rather than overfitting to rigid scalars.

\textbf{Interpretability and Latent Structure of the Physics Descriptor $\zp$.} 
We analyze whether the physics descriptor learns structured representations of the test-scene physics. Specifically, in Push-T, we collect 10 interaction clips in each of the 8 unseen center of mass (CoM) locations ($P_1$--$P_8$) and encode each clip into a physics descriptor token using the context encoder $g_\phi$. We then compute the pairwise cosine similarities between these tokens and average them across each CoM pair.  
As shown in \autoref{fig:8-ctx-token}, descriptors from adjacent CoM locations exhibit strong positive correlation, while CoMs that are far apart yield negative similarity values.
This indicates that our training objective enables the context encoder to capture meaningful physics properties correlated with the scene dynamics, even without using ground-truth physics labels during training.

%% file: sections/discussion.tex
\section{Discussion}

\textbf{Limitations.}
Our approach has several limitations. 
First, training \method requires grouping trajectories by the physics configurations under which they were collected. While partitioning data by environment identity is far less demanding than deriving analytical equations, curating grouped data from in-the-wild video remains non-trivial. One potential direction to relax this is leveraging self-supervised representation learning to automatically 
cluster latent transition dynamics into environment groups.
Moreover, the proposed test-time calibration depends on the quality of the interaction prompt given to the model. While we find that simple playful or even random policies often suffice, future work could explore active or curiosity-driven exploration to collect more informative prompts.
Finally, our framework assumes the physics properties are stationary: the dynamics remain unchanged within an episode. Extending this approach to non-stationary environments (e.g., fluctuating wind fields) may require additional techniques such as reactive prompting or online descriptor updates.

\textbf{Conclusion.}
We believe that world models should go beyond generic visual fidelity toward instance-specific physics-grounded prediction. 
In this work, we take a step toward this goal by introducing an in-context calibration framework that grounds action-conditioned world models in the actual physics dynamics of novel test scenes. 
We achieve this by prompting the model with an interaction sequence collected within the test scene, enabling zero-shot calibration without finetuning or analytical physics priors. 
Evaluations across diverse environments demonstrate that our approach substantially improves predictive accuracy and downstream visual planning.

%% file: sections/acknowledgement.tex
\subsubsection*{Acknowledgments}
This work is supported in part by the National Science Foundation under CAREER Award No. 2544116.

%% file: sections/appendix.tex
\newpage
\appendix
\setcounter{figure}{0}
\renewcommand{\thefigure}{A\arabic{figure}}  %
\setcounter{table}{0}
\renewcommand{\thetable}{A\arabic{table}}

{
\begingroup
\hypersetup{linkcolor=black}
\setstretch{0.7}
\tableofcontents
\endgroup
}

\newpage

\section*{Appendix}
In this appendix, we first present our implementation details in Appendix~\ref{sec:appendix_implementation}, such as description of each baseline, model architecture of \method, hyperparameters, as well as our training and evaluation setups. 
Next, in Appendix~\ref{sec:appendix_environments}, we provide detailed description of each environment we used along with environment-specific implementation. 
Following this, we include additional quantitative results in Appendix~\ref{sec:appendix_additional_results}, such as comparison of multiple visual metrics on next-frame prediction and analyses of training and inference speed.
Finally, we append additional qualitative results in Appendix \ref{sec:appendix_qualitative_results}.

\section{Implementation Details}
\label{sec:appendix_implementation}

\subsection{Baselines} \label{sec:appendix_baselines}
In this section, we provide a detailed description for each baseline we used in our paper.
Please see the description of each method below:
\begin{enumerate}[label=\alph*), leftmargin=*]
    \item \textbf{DINO-WM}~\citep{zhou2025dino}: 
    We follow the implementation in the official codebase, using a frozen DINOv2~\citep{oquab2023dinov2} backbone as the visual encoder $E_\text{v}$. 
    The input to the dynamics predictor $f_\theta$ is an observation-action sequence of $K=3$ steps, including $K-1$ historical transitions alongside the current observation and action.
    The predictor predicts one-step future observation $\hat{o}_{t+1}$.
    Because this baseline receives no in-context interaction prompt, it must infer transition dynamics solely from this local history window. 
    We observe that DINO-WM exhibits severe rollout drift when physical properties vary across test scenes.

    \item \textbf{DINO-WM + Context Prefix:} This baseline uses the same model architecture and training objective as DINO-WM. 
    The only difference is that it additionally  
    feeds the latent embeddings of the whole in-context interaction prompt to the dynamics predictor $f_\theta$. 
    As a result, for this baseline, the total input window size of $f_\theta$ is $T_c + K$.
    Since transformer self-attention scales quadratically with the number of input tokens, this uncompressed prefix adds substantially more computation to both model training and inference (see \autoref{fig:7-inference-time-main} and \autoref{tab:pusht_inference_efficiency} for runtime comparisons).

    \item \textbf{DINO-WM + Long History:}
    This baseline uses the same model architecture and training objective as DINO-WM. 
    The only difference is that it extends the target observation history from 3 to 15 consecutive frames (\textit{i.e.,} $K=15$). The goal of this baseline is to study whether extending history can enable effective physics-calibration in world models.
    
    \item \textbf{LeWM}~\citep{maes2026leworldmodel}: 
    LeWM is a latent world model directly trained from scratch (without using frozen pre-trained models like DINOv2), hence it learns task-specific latent visual representations and action-conditioned dynamics. We follow the implementation in its official codebase. 
    Based on LeWM, we also include \textbf{LeWM + \shortmethod (Ours)}, where we integrate 
    the proposed physical calibration module into LeWM, to show that our proposed physics calibration method is a plug-and-play module that can potentially extend to any latent world models.
    
    \item \textbf{TC-LeWM}~\citep{liu2026tcsigreg}: 
    This method applies SIGReg to temporally centered residuals rather than the full latent distribution, preventing the suppression of fine-grained moving components.
    We include this to enhance the LeWM baseline, as vanilla LeWM's latent representations fail to capture the ball in the BeerPong environment (yielding very low IoU and planning success). Similarly, we include \textbf{TC-LeWM + \shortmethod (Ours)} to evaluate whether our proposed method can be applied to this improved LeWM baseline.

    \item \textbf{Temporal Straightening (TS-WM)}~\citep{wang2026temporal}: 
     we adopt this TS-WM variant: DINOv2 (patch) + proj configuration with $16 \times 16 \times 8$ spatial features, as it achieves the strongest overall performance reported in the original paper.
    Similar to DINO-WM, 
    this variant uses a frozen DINOv2 visual encoder and a causal transformer dynamics predictor,
    but inserts an additional lightweight projection head between the visual encoder and the predictor.
    During training, TS-WM introduces a curvature regularizer to enforce a more structured latent representation, and such regularization has demonstrated stronger planning performance. 
    We include this baseline to study whether additional regularization can promote physics grounding in world models.

    \item \textbf{DINO-WM + Estimated Physics$^\dagger$:} This baseline first estimates the physics configurations of the scenes (e.g., the center of mass in Push-T or the joint damping in Reacher) from a context clip and feeds the estimated value as condition for the dynamics predictor $f_\theta$.
Specifically, we train a separate physics estimator based on the same context-encoder architecture as our method but use the ground-truth physics labels as supervision. 
At inference time, the estimated parameters are provided to the dynamics predictor as additional input for conditioning. 
This baseline compares explicit, supervised physics estimation with our end-to-end in-context calibration method. Note that this baseline requires predefined physics labels for training, while our method does not.

    \item \textbf{DINO-WM + Privileged Physics$^\ddagger$:} 
    This baseline is based on DINO-WM but can access the ground-truth physics configurations of the scenes during both training and inference. 
    Specifically, the dynamics predictor directly receives these true physical parameters as supplementary inputs alongside standard observation and action tokens.

\end{enumerate}

\subsection{Implementation of \method}

\paragraph{Software and Hardware.}
Our experiments are implemented in Python 3.10 on Red Hat Enterprise Linux 8.10.
We train and evaluate our models using NVIDIA L40S GPUs and NVIDIA RTX PRO 6000 Blackwell GPUs interchangeably.

\paragraph{Base World Models.}
We implemented \method on two latent world-model backbones, DINO-WM~\citep{zhou2025dino} and LeWM~\citep{maes2026leworldmodel}, denoted as DINO-WM+\shortmethod and LeWM+\shortmethod, as in \autoref{tab:main_pusht_reacher_results} and \autoref{tab:main_beerpong_fetchslide_results}.
The base architecture and optimization hyperparameters used for the two backbones are summarized in \autoref{tab:app_wm_hyperparameters}.

\begin{table}[htbp]
    \centering
    \small
    \setlength{\tabcolsep}{8pt}
    \adjustbox{max width=\textwidth}{
    \begin{tabular}{lcc}
        \toprule
        \textbf{Configuration} & \textbf{DINO-WM-based models} & \textbf{LeWM-based models} \\
        \midrule
        Vision Encoder $E_\text{v}$ & Frozen DINOv2 (ViT-Small/14) & Learnable ViT-Tiny/14 \\
        Visual Token Dimension $d_z$ & 384 & 192 \\
        Visual Tokens per Frame & $196$ ($14 \times 14$) & $1$ \\
        Dynamics Predictor $f_\theta$ & Causal Transformer & Causal Transformer \\
        \quad Layers & 6 & 6 \\
        \quad Attention Heads & 16 & 16 \\
        FFN Hidden Dimension $d_{\mathrm{mlp}}$ & $2048$ & $2048$ \\
        Predictor Dropout & $0.1$ & $0.1$ \\
        Optimizer & AdamW & AdamW \\
        Model Learning Rate & $1 \times 10^{-4}$ & $5 \times 10^{-5}$ \\
        Weight Decay & $0.01$ & $1 \times 10^{-3}$ \\
        Gradient Clipping Norm & $1.0$ & $1.0$ \\
        Global Batch Size & $48$ & $128$ \\
        Observation History Length $K$ & $3$ steps & $3$ steps \\
        Proprioceptive Loss Weight & $0.5$ & $0.5$ \\
        \bottomrule
    \end{tabular}
    }
    \vspace{-3pt}
    \caption{\textbf{Model Architectures and Training Hyperparameters.} The base configurations used by the DINO-WM and LeWM model families, corresponding to results shown in \autoref{tab:main_pusht_reacher_results}
    and \autoref{tab:main_beerpong_fetchslide_results}.
    }
    \label{tab:app_wm_hyperparameters}
    \vspace{-5pt}
\end{table}

\paragraph{Context Encoder.}
For all our experiments, the context encoder $\ctxenc$ maps a 12-frame ($T_c=12$) interaction sequence
to a 64-dimensional physics descriptor \zp.
Each frame is embedded by a ResNet-based visual encoder $E_c^v$ into 64 two-dimensional keypoints via spatial softmax~\citep{finn2016deep}.
The world model's input action is embedded by a single linear layer
$E_c^a:\mathbb{R}^{N_a \times d_a}\rightarrow\mathbb{R}^{10}$, where $N_a$ denotes the length of the input action chunk and $d_a$ denotes the dimension of a single action.
For Push-T, the 4D observable pusher state is additionally embedded by
$E_c^p:\mathbb{R}^{4}\rightarrow\mathbb{R}^{10}$.
The resulting per-step features are concatenated channel-wise and mapped to one 512-dimensional latent embedding (i.e., data of each timestep is now represented by one latent embedding).
A four-layer Transformer encoder with eight attention heads takes latent embeddings of all $T_c$  timesteps as input and pools them with a learnable token $\mathbf{z}_\text{cls}$.
Finally, the output corresponding to $\mathbf{z}_\text{cls}$ is
fed to an MLP
, whose output is the \textit{physics descriptor token} \zp. For simplicity, we set the dimension of \zp to 64 across all our experiments.
We summarize the hyperparameters of the context encoder $g_\phi$ in \autoref{tab:app_ctx_encoder_specs}.

\begin{table}[htbp]
    \centering
    \small
    \setlength{\tabcolsep}{12pt}
    \begin{tabular}{lc}
        \toprule
        \textbf{Hyperparameter} & \textbf{Value} \\
        \midrule
        ResNet GroupNorm Groups & 8 \\
        Spatial Projection Channels & 64 \\
        Number of Spatial Keypoints & 64 \\
        Visual Feature Dimension & 128 \\
        Action Embedding Dimension & 10 \\
        Proprioceptive Embedding Dimension & 10 \\
        Temporal Encoder Depth & 4 \\
        Attention Heads & 8 \\
        Hidden Dimension & 512 \\
        FFN Hidden Dimension & 1024 \\
        Physics Descriptor Dimension & 64 \\
        Context Length $T_c$ & 12 frames \\
        Learning Rate & Same as the base world model \\
        \bottomrule
    \end{tabular}
    \caption{\textbf{Hyperparameters of the Context Encoder $\ctxenc$.}
     We use this set of hyperparameters for all our experiments. Note that similar to DINO-WM~\citep{zhou2025dino}, we apply a frameskip for the input context clip, such that the duration of the 12-frame context spans several seconds in the environments. 
     See \autoref{tab:environment_summary} for per-environment setup.}
    \label{tab:app_ctx_encoder_specs}
\end{table}

\paragraph{Context Sampling and Backbone Integration.}
For each prediction segment sampled from a physical-configuration group $\mathcal{G}_j$, we independently sample the interaction context from a different episode in the same group.
Thus, the context shares the same underlying physical dynamics as the prediction target without containing trajectory-specific future observations.
All contexts contain $T_c=12$ observations sampled at the  model prediction gap of the corresponding world model.

For Push-T and Reacher, we uniformly sample the $T_c=12$-frame context window within a prompt episode. That 12-frame clip typically spans a few seconds.
For BeerPong, we sample the context from the beginning of the episode, such that the context contains both the initial release status and ball-table interaction that reveals the contact damping.
Similarly, for FetchSlide, we also start the context from the beginning of each episode to include the initial robot-puck interaction sequence and also because episodes in FetchSlide are relatively short.
If the in-context prompt episode has fewer than $12$ frames, we pad the context to length $T_c$ by repeating the final available frame.

For DINO-WM+\shortmethod, \zp is broadcast across all spatial patches and temporal steps and concatenated with the predictor
features along the channel dimension.
For LeWM+\shortmethod, the 64-dimensional descriptor is concatenated with the
192-dimensional per-step latent representation before feeding into the causal future predictor $f_\theta$, while action conditioning follows the original LeWM implementation through AdaLN-zero.

\paragraph{Implementation of the Temporal Straightening (TS-WM) Baseline.}
For Temporal Straightening (TS-WM)~\citep{wang2026temporal}, we follow the original implementation and use the DINOv2 (patch) + projection configuration with the default curvature regularization and optimization settings given by the authors, as it is the best-performing model reported in the original paper.
We summarize the corresponding hyperparameters in \autoref{tab:app_ts_specs}.

\begin{table}[htbp]
    \centering
    \small
    \setlength{\tabcolsep}{12pt}
    \begin{tabular}{lc}
        \toprule
        \textbf{Hyperparameter} & \textbf{Value} \\
        \midrule
        Projection Intermediate Channels & 192 \\
        Projection Output Channels & 8 \\
        Projected Feature Map Shape & $16 \times 16 \times 8$ \\
        Curvature Loss Weight $\lambda_{\mathrm{curv}}$ & $0.1$ \\
        Projection Head Learning Rate & $1 \times 10^{-5}$ \\
        Predictor Learning Rate & $5 \times 10^{-4}$ \\
        \bottomrule
    \end{tabular}
    \caption{\textbf{Hyperparameters of the Temporal Straightening Baseline (TS-WM).} We use the 
    DINOv2 (patch) + projection variant and follow the hyperparameters used in the original paper~\citep{wang2026temporal}.}
    \label{tab:app_ts_specs}
\end{table}

\input{sections/pseudo_code}

\input{sections/cem_pseudo_code}

\clearpage

\section{Environment Training and Evaluation Details}
\label{sec:appendix_environments}
In this section, we provide detailed descriptions of our experiments in each environment, including environment setup, training and evaluation details. 
\autoref{tab:environment_summary} summarizes the dataset scale and hyperparameters of the world models in each environment. The resolution of the image observation is  $224\times224$ for all four environments.

\begin{table}[h]
    \centering
    \small
    \setlength{\tabcolsep}{2.5pt}
    \renewcommand{\arraystretch}{1.12}
    \adjustbox{max width=\textwidth}{
    \begin{tabular}{lcccc}
        \toprule
        & \textbf{Push-T} & \textbf{Reacher} & \textbf{FetchSlide} & \textbf{BeerPong} \\
        \midrule
       
        Training / evaluation physical configurations & 135 / 15 & 135 / 15 & 135 / 15 & 135 / 15 \\
        Episodes per configuration & 101 & 20 & 240 & 60 \\
        Total dataset episodes & 15,150 & 3,000 & 36,000 & 9,000 \\
        
        Control interval (s) & 0.10 & 0.04 & 0.04 & 0.01 \\
        world model \# of frameskip & 5 & 5 & 5 & 15 \\
        world model per-step interval (s) & 0.50 & 0.20 & 0.20 & 0.15 \\
      
        Environment action dimension & 2 & 2 & 2 & 2 \\
        World-model action input dimension & 10 & 10 & 8 & 2 \\
        \bottomrule
    \end{tabular}
    }
    \caption{\textbf{Task and Dataset Summary.} All environments use $224\times224$ RGB observations.}
    \label{tab:environment_summary}
\end{table}

\subsection{Push-T}
\label{sec:appendix_pusht}

\paragraph{Task and Physics Variation.}
Push-T is a planar manipulation task in a $512\times512$ pixel workspace, where a circular pusher of radius 15 pixels moves a rigid T-shaped block toward a goal configuration specified by an RGB image. The block consists of a $120\times30$ pixel crossbar and a $30\times90$ pixel stem. We uniformly sample the block's local center of mass (CoM) inside its T-shaped interior and hold it fixed within each episode. Changing the CoM changes the moment arms of contact forces and the resulting translational and rotational response, while the block appearance remains fixed.

\paragraph{Dataset and Action Space.}
The dataset contains 150 CoM configurations, with 135 used for training and 15 held out for evaluation, totaling 15,150 episodes with 101 episodes per configuration. 
We train a CoM-agnostic PPO policy that targets pushing the T-block to the light-green area and use this policy to collect trajectories for training, since a random policy cannot ensure sufficient and meaningful interaction. 
In theory, any policy that produces sufficient contacts can serve as the data-collection policy. 
Each raw action specifies an absolute 2D pusher target in simulator pixel coordinates, tracked by a PD controller at a control frequency of 10 Hz.
Following \cite{zhou2025dino}, we inject independent Gaussian noise into the output action during policy execution. The noise uses ten different standard deviations: $\sigma\in\{0.1,0.3,0.5,1.0,1.5,2.0,2.5,3.0,7.0,15.0\}$ pixels. 
This online noise injection enhances dataset coverage and generalization, which is particularly essential for downstream planning using CEM, as its initial optimization phase relies heavily on exploring noisy action trajectories.
We subsample observations every 5 control steps and concatenate the 5 2D actions in between into a stacked 10D action. We use this 10D action as the input to the world models, yielding a prediction interval of 0.5 seconds.
Models receive $224\times224$ RGB images and a 4D pusher proprioceptive input $(x,y,v_x,v_y)$, whose velocity entries are zero-filled so that only poisition information is provided.

\paragraph{Training and evaluation.}
Please see \autoref{tab:app_wm_hyperparameters}, \autoref{tab:app_ctx_encoder_specs}, and \autoref{tab:app_ts_specs} for detailed model hyperparameters, and \autoref{tab:training_time_pusht_reacher} for training wall-clock durations.
Visual prediction IoU is computed on the union of the agent and block foreground masks. Each planning task pairs a single current observation with a goal observation recorded 15 control steps later in the same trajectory. Success requires the Euclidean position error over the concatenated agent and block coordinates to be below 20 pixels and the wrapped block-orientation error to be below $\pi/9$ ($20^\circ$). 
For open-loop execution, we set the planning horizon to 25 control steps (i.e., 5-step future prediction in the world model). 
In \autoref{tab:main_pusht_reacher_results}, we report the success rate under this fixed 25-step budget when directly executing the planned action sequence from the initial observation.
To leverage the world model for dynamic planning, we also experiment with model predictive control (MPC), in which the model actively replans if the task has not succeeded. 
\autoref{tab:mpc_success} evaluates cumulative success rates with at most two replans (i.e., budgets of 25, 50, and 75 control steps).
We observe that the success rate of our method consistently increases with larger step budgets, as iterative replanning allows for continuous error correction.

\subsection{Reacher}
\label{sec:appendix_reacher}

\paragraph{Task and physical variation.}
We modify the Reacher environment from the DeepMind Control Suite~\citep{tassa2018deepmind}, where a planar two-link arm moves toward a goal configuration specified by an RGB image. The proximal and distal links have lengths of 0.12 and 0.10 meters, respectively. We vary the two joint damping coefficients over $[0.005,0.020]\,\mathrm{N\cdot m\cdot s/rad}$ using Latin-hypercube sampling and hold them fixed within each episode. Changing joint damping alters the two-link arm's resistance to motion without changing the visual appearance.

\paragraph{Dataset and action space.}
The dataset contains 150 joint-damping configurations, with 135 used for training and 15 held out for evaluation, totaling 3,000 episodes with 20 episodes per configuration. Each episode contains 100 control steps and starts from randomly sampled joint angles with zero initial joint velocities.
Unlike Push-T, random torque commands naturally induce diverse joint motions and provide broad coverage of the joint-state and velocity space, so no task-specific data-collection policy is required.
Each 2D action is sampled independently from $[-1,1]^2$ and specifies normalized motor commands corresponding to joint torques in $[-0.05,0.05]\,\mathrm{N\cdot m}$. Each command is held for two 0.02-second simulation steps, giving a control frequency of 25 Hz. Similar to Push-T, we subsample observations every 5 control steps and concatenate the five intervening 2D actions into a 10D world-model action, yielding a prediction interval of 0.2 seconds. 
Unlike Push-T, Reacher does not provide proprioception inputs.

\paragraph{Training and evaluation.}
Visual prediction IoU is computed on the union of the two arm links' foreground masks.
Each planning task pairs a single current observation with a goal observation recorded 25 control steps later in the same trajectory. 
Success requires the angular error of each joint to be below $10^\circ$ ($\pi/18$ radians).
Similar to Push-T, we use a 25-control-step planning horizon (i.e., 5-step future prediction in the world model), and report the 25-step budget planning result in \autoref{tab:main_pusht_reacher_results}. We additionally evaluate MPC with at most two replans, yielding cumulative budgets of 25, 50, and 75 control steps, as reported in \autoref{tab:mpc_success}.
Note that repeated replanning substantially reduces the effect of accumulated model error in Reacher, allowing even imperfect models to iteratively approach the target and resulting in high MPC success rates.
We observed that LeWM-based models perform particularly well on Reacher, which may be attributed to its encoder being task-specific and able to focus on essential structures such as the angles of the joints, whereas DINOv2 is trained as a general-purpose image encoder.

\subsection{BeerPong}
\label{sec:appendix_beerpong}

\paragraph{Task and physical variation.}

We modify the BeerPong environment from Fancy Gym~\citep{fancy_gym}, where a simulated Barrett Whole-Arm Manipulator (WAM) throws a ball toward a cup across a table.
We omit an explicit gripper and instead kinematically attach the ball to the end effector during the first five control steps, after which it is released with the end effector's instantaneous velocity. The robot starts from a fixed configuration, while the cup position is sampled along the table's central axis and remains fixed within each episode. We uniformly sample the ball-contact damping parameter over $[10,40),\mathrm{N\cdot s/m}$ and hold it fixed within each episode.
This changes energy dissipation during contact and the resulting bounce trajectory without changing visual appearance.
We empirically choose the contact-damping range as lower values produce insufficient rebound while higher values lead to excessively unstable ball trajectories. With damping parameter inside this range, most episodes are physically feasible, with some throws capable of making the ball reach the cup after one or two bounces on the table.
The table measures approximately $0.90\times1.68$ meters, the cup has an inner radius of 0.10 meters, and the ball has a radius of 0.049 meters and a mass of 0.10 kilograms.

\paragraph{Dataset and action space.}
The dataset contains 150 contact-damping configurations, with 135 used for training and 15 held out for evaluation, totaling 9,000 episodes with 60 episodes per configuration. Each episode contains 300 control steps. 
A 2D throw command is sampled uniformly from $[2.5,3.0)\,\mathrm{m/s}$ for forward end-effector velocity and $[6.0,8.0)\,\mathrm{m/s}$ for upward end-effector velocity.
After releasing the ball, the robot arm retracts to a fixed configuration to prevent its post-throw pose from leaking information about the throw command.
Each control step spans two 0.005-second simulation steps, giving a control frequency of 100 Hz. We subsample observations every 15 control steps, yielding a prediction interval of 0.15 seconds (the time gap between two predicted frames by the world model). 
Since the robot actions cannot affect the ball after the robot releases the ball, we provide the normalized 2D throw command as the world model's action input at every prediction step, both before and after the release. 
The world models receive $224\times224$ RGB images without proprioception inputs.

\paragraph{Training and evaluation.}
Visual prediction IoU is computed on the ball's foreground mask. Each planning task consists of a single initial observation and a final observation of a successful in-cup throw (i.e., an image in which the ball is inside the cup). 
The success criterion is that the ball can be thrown inside the cup and stay in the cup.
We evaluate each method on 75 shared targets, where all planning tasks share the same initial robot and ball configuration and differ in the randomly sampled cup position and the table contact damping of the held-out configuration. 
Different from Push-T and Reacher, BeerPong is a one-shot planning task, in which the planner optimizes a single 2D throw command.

\subsection{FetchSlide}
\label{sec:appendix_fetchslide}

\paragraph{Task and Physics Variation.}
We modify the FetchSlide-v4 environment from Gymnasium Robotics~\citep{gymnasium_robotics2023github}, where a Fetch manipulator strikes a cylindrical puck toward a goal position on a table. The puck has a radius of 0.05 meters and a height of 0.04 meters.
We independently sample puck mass $m\in[0.04,0.36)\,\mathrm{kg}$, surface friction $\mu\in[0.0305,0.0435)$, and fluid density $\rho\in[0,40)\,\mathrm{kg/m^3}$ from uniform distributions and hold them fixed within each episode. These numbers are selected under the principles that the environments can remain valid. For example, if we set the upper limit of the surface friction to a very large value, in most of the cases, the puck might simply stop very quickly, resulting in tiny difference between different environments.

At a high level, the interplay between these physics properties can be described as follows.
For the same strike action, lower puck mass results in a higher post-strike velocity. 
The subsequent puck's deceleration depends jointly on surface friction and aerodynamic drag, where stronger friction and higher fluid density will increase deceleration rate. 
In addition, since we do not vary the shape of the puck, given the same air density, a puck of larger mass will decelerate slower compared to a lighter puck.

\paragraph{Dataset and action space.}
The dataset contains 150 physical configurations, with 135 used for training and 15 held out for evaluation, totaling 36,000 episodes with 240 episodes per configuration. Each strike is specified by a 2D action sampled uniformly from $[0.35,0.85)\times[-0.8,0.8)$.

Following the original setting of FetchSlide-v4 \citep{gymnasium_robotics2023github}, action is expanded into a 4D Cartesian command $(a_x,a_y,0,0)$, where the first three components specify end-effector displacement and the final component controls the gripper, which is always closed. The same command is applied for the first two control steps. Each control step spans 0.04 seconds. We subsample observations every five control steps, yielding a prediction interval of 0.2 seconds, and concatenate the two initial 4D commands into a normalized 8D condition provided at every world-model step.
After striking the puck, the robot arm moves to a fixed position to prevent information leakage from the post-strike pose.
Models receive $224\times224$ RGB images without proprioception input.

\paragraph{Training and evaluation.}
Visual prediction IoU is computed on the puck's foreground mask.
All planning tasks share the same initial robot and puck configuration and differ only in the target puck position.
Similar to BeerPong, FetchSlide is a one-shot planning task, where the planner optimizes a single 2D strike command, after which the puck will move on the tabletop until it stops, without additional intervention from the robot. 
We evaluate all methods across a shared benchmark of 75 planning targets, defining the success of one target when the final puck position lies within $0.03\text{ m}$ Euclidean distance of the target position on the table.

\clearpage
\section{Additional Quantitative Results}
\label{sec:appendix_additional_results}
We first present performance comparison of next-frame prediction in Section~\ref{sec:appendix_next_frame}, where we conduct teacher-forcing style evaluation (only score the immediate next frame when computing the metrics), complementing the open-loop rollout evaluation presented in the main paper. 
Then, we show the planning task success rates over more replanning budgets in Section \ref{sec:appendix_mpc}.
Following this, we present the training time and inference time comparison of each method in Section \ref{sec:appendix_training_time} and Section \ref{sec:appendix_inference_time}, respectively.
Finally, we report how different success thresholds will affect the planning success rate in Section \ref{sec:appendix_success_thresholds}.

\subsection{Next-Frame Prediction Performance}\label{sec:appendix_next_frame}

We provide additional next-frame prediction results on all four environments in \autoref{tab:next_frame_four_envs}. We report PSNR, SSIM, LPIPS, and foreground IoU to evaluate both pixel-level reconstruction quality and foreground prediction accuracy. Across all four environments, incorporating PhysiCal consistently improves the next-frame prediction performance of both DINO-WM and LeWM.

\input{tables/table_next_frame_four_envs}

\subsection{MPC Results Across Execution Budgets}
\label{sec:appendix_mpc}

Following DINO-WM \citep{zhou2025dino}, in addition to the open-loop CEM planning, we also evaluate model predictive control (MPC), where the model replans from the newly observed state after executing a sequence of optimized actions.
Specifically, in Push-T and Reacher, CEM optimizes an action sequence spanning 25 control steps (i.e., 5 world model steps due to action stacking) using the world model. The full sequence is then executed in the environment. 
If this action sequence cannot complete the task, we replan an additional 25 control steps. We allow at most two additional replans, corresponding to maximum 25, 50, and 75 control steps.

The result is shown in \autoref{tab:mpc_success}. +PhysiCal consistently improves performance on Push-T under all maximum numbers of control steps, showing that the performance of our method can scale with additional compute. 
In Reacher, +PhysiCal shows a strong advantage over context-free baselines after the first 25 execution steps. 
However, the gap narrows after replanning because the task is highly feedback-correctable and the structure of the task is relatively easy (only consisting of two rotary joints), allowing MPC to quickly compensate for systematic model bias.

\begin{table}[H]
    \centering
    \small
    \setlength{\tabcolsep}{5pt}
    \begin{tabular}{@{}l c c c@{\hspace{14pt}}c c c@{}}
         & \multicolumn{3}{c}{\textbf{Push-T}} & \multicolumn{3}{c}{\textbf{Reacher}} \\
        \cmidrule(lr){2-4}\cmidrule(lr){5-7}
        \textbf{Method}  & 25 & 50 & 75 & 25 & 50 & 75 \\
        \midrule
        TS-WM & 53.3 & 54.7 & 56.0 & 41.3 & 65.3 & 68.0 \\
        DINO-WM & 53.3 & 54.7 & 56.0 & 52.0 & 76.0 & 82.7 \\
        ~ + Long History & 54.7 & 56.0 & 58.7 & 56.0 & \underline{92.0} & \underline{96.0} \\
        ~ + Context Prefix & 58.7 & 60.0 & 60.0 & 61.3 & 84.0 & 92.0 \\
        ~ + Estimated Physics$^\dagger$ & 57.3 & \underline{62.7} & \underline{65.3} & \underline{69.3} & 85.3 & 90.7 \\
        ~ + Privileged Physics$^\ddagger$ & \underline{60.0} & \underline{62.7} & 64.0 & \textbf{86.7} & \textbf{100.0} & \textbf{100.0} \\
        \rowcolor{green!17}
        ~ + \shortmethod (Ours) & \textbf{62.7} & \textbf{66.7} & \textbf{68.0} & 61.3 & 80.0 & 93.3 \\
        \midrule
        LeWM & \underline{48.0} & \underline{48.0} & \underline{49.3} & \underline{60.0} & \underline{84.0} & \underline{90.7} \\
        \rowcolor{green!17}
        ~ + \shortmethod (Ours) & \textbf{61.3} & \textbf{62.7} & \textbf{62.7} & \textbf{64.0} & \textbf{86.7} & \textbf{92.0} \\
        \bottomrule
    \end{tabular}
    \caption{\textbf{MPC success rates on Push-T and Reacher under different maximum numbers of control steps.} Max. control steps denotes the maximum number of control steps allowed across all replanning iterations. Incorporating PhysiCal substantially improves both DINO-WM- and LeWM-based models across all allowed numbers of replans.}
    \label{tab:mpc_success}
    \par\smallskip
\end{table}

\subsection{Model Training Efficiency}
\label{sec:appendix_training_time}
We report  the training time for all world models evaluated in this work, comparing the number of epochs, GPU configurations, and wall-clock training hours across tasks. 
We use two types of GPUs: L40S denotes NVIDIA L40S, and PRO 6000 denotes NVIDIA RTX PRO 6000.

As shown in \autoref{tab:training_time_pusht_reacher}, for DINO-WM based models, though our method introduces around 11\% additional training cost over the vanilla DINO-WM on Push-T,
the training speed of +PhysiCal remains faster compared to the Long History and Context Prefix models. 
Due to the significantly higher training time for the Long History and Context Prefix models, for fair comparison, we align their GPU hours spent on training with other methods. 
Specifically, for Long History we trained only 5 epochs in Push-T, and for Context Prefix we trained only 5 epochs in Push-T, 12 epochs in Reacher and 3 epochs in BeerPong.

\begin{table}[htbp]
    \centering
    \small
    \setlength{\tabcolsep}{3pt}
    \adjustbox{max width=\textwidth}{
    \begin{tabular}{l c c r c c r}
        & \multicolumn{3}{c}{\textbf{Push-T}} & \multicolumn{3}{c}{\textbf{Reacher}} \\
        \cmidrule(lr){2-4} \cmidrule(lr){5-7}
        \textbf{Method} & Epochs & GPUs & Time (hours) & Epochs & GPUs & Time (hours) \\
        \midrule
        TS-WM & 10 & 4 $\times$ L40S & 25 & 20 & 4 $\times$ L40S & 9 \\
        DINO-WM & 10 & 4 $\times$ L40S & 37 & 20 & 4 $\times$ L40S & 13 \\
        ~ + Long History & 5 & 4 $\times$ L40S & 49 & 20 & 4 $\times$ L40S & 26 \\
        ~ + Context Prefix & 5 & 4 $\times$ L40S & 52 & 12 & 4 $\times$ L40S & 22 \\
        ~ + Privileged Physics$^\ddagger$ & 10 & 4 $\times$ L40S & 37 & 20 & 4 $\times$ L40S & 13 \\
        \rowcolor{green!17}
        ~ + \shortmethod (Ours) & 10 & 4 $\times$ L40S & 41 & 20 & 4 $\times$ L40S & 20 \\
        \midrule
        LeWM & 10 & 4 $\times$ L40S & 25 & 20 & 4 $\times$ L40S & 5 \\
        \rowcolor{green!17}
        ~ + \shortmethod (Ours) & 10 & 4 $\times$ L40S & 37 & 20 & 4 $\times$ L40S & 14 \\
        \bottomrule
    \end{tabular}
    }
    \vspace{-6pt}
    \caption{\textbf{Training time of each world model in Push-T and Reacher.} 
    For fair comparison, we approximately align the model training GPU hours of each method. Due to the significantly longer per-iteration time, we train +Long History and +Context Prefix for less epochs. 
    Numbers are for general reference only, as we notice that the training speed (even of the same model) can slightly vary in our cluster, which is potentially due to disk loading speeds, variance across different GPU nodes, network interconnect congestion, etc.}
    \label{tab:training_time_pusht_reacher}
\end{table}

\vspace{-10pt}
\subsection{Model Inference Efficiency}
\label{sec:appendix_inference_time}

We measure latent model inference on Push-T using one NVIDIA RTX PRO 6000 Blackwell Server Edition GPU and FP32 with TF32 disabled.
For next-frame prediction, we use 15 observed frames for Long History ($K=15$) and 3 for the remaining methods ($K=3$).
In open-loop rollouts, each model starts with three ground-truth observation frames. Long History grows its history autoregressively up to its 15-frame window.
Context Prefix, Estimated Physics, and our method use the same 12-frame observation-action sequence as the context prompt.
For planning inference time, we use the same CEM settings as in our main experiments and report the mean of ten repeated CEM solves.

\autoref{tab:pusht_inference_efficiency} shows that our method incurs a 2.30\,ms condition-encoding cost. Including this cost, next-frame prediction takes 7.82\,ms, adding 2.24\,ms over DINO-WM (5.58\,ms), compared with 13.91\,ms for both Context Prefix and Long History.
For open-loop rollouts, our method requires 16.03\,ms for 5-step prediction and 45.83\,ms for 20-step prediction, remaining close to DINO-WM (13.78\,ms and 44.48\,ms, respectively), compared with Context Prefix (48.36\,ms and 177.61\,ms) and Long History (18.05\,ms and 131.27\,ms).
These results indicate that compressing context into a compact representation preserves near-baseline inference cost, avoiding the computational penalties of uncompressed prefixes or long observation histories.

\begin{table}[htbp]
    \centering
    \small
    \setlength{\tabcolsep}{5pt}
    \adjustbox{max width=\textwidth}{
    \begin{tabular}{l c c c c c c}
        \textbf{Method} & \shortstack{Condition Encoding\\(ms)} & \shortstack{Next-Frame\\(ms)} & \shortstack{Rollout-5f\\(ms)} & \shortstack{Rollout-10f\\(ms)} & \shortstack{Rollout-20f\\(ms)} & \shortstack{Planning\\(s)} \\
        \midrule
        DINO-WM & -- & 5.58 & 13.78 & 24.01 & 44.48 & 32.66 \\
        ~ + Context Prefix & 4.89 & 13.91 & 48.36 & 91.44 & 177.61 & 323.09 \\
        ~ + Long History (15 frames) & -- & 13.91 & 18.05 & 48.97 & 131.27 & 43.96 \\
        ~ + Privileged Physics$^\ddagger$ & -- & 5.56 & 13.67 & 23.82 & 44.05 & 32.55 \\
        ~ + Estimated Physics$^\dagger$ & 2.31 & 7.81 & 16.06 & 26.36 & 46.97 & 32.61 \\
        \rowcolor{green!17}
        ~ + \shortmethod (Ours) & 2.30 & 7.82 & 16.03 & 26.31 & 45.83 & 32.67 \\
        \bottomrule
    \end{tabular}
    }
    \vspace{-7pt}
    \caption{\textbf{Push-T model inference latency.}
(1) \textbf{Condition encoding}: Measures the latency to encode the 12 conditioning frames into model representations. Our method uses the lightweight context encoder, Context Prefix uses the primary vision backbone (i.e., DINOv2), and Estimated Physics uses a context encoder followed by a physical-parameter regressor. Methods without context conditioning do not require this step.
    (2) \textbf{Next-Frame Prediction (Teacher Forcing)}: Measures the latency to encode the real observation history (3 frames for standard models, 15 frames for Long History) and, where applicable, a 12-frame context clip, and predict one next latent frame under recorded actions.
    (3) \textbf{Open-Loop Rollout (5, 10, and 20 frames)}: Measures the total latency of predicting $H$ future latent frames ($H \in \{5, 10, 20\}$) from three unencoded observed frames and, where applicable, a 12-frame unencoded context clip. Long History retains its 15-frame capacity and fills its history autoregressively.
    (4) \textbf{Planning}: Measures the total latency of one CEM solve over 25 control steps (5 model steps), using 300 candidate action sequences, 30 elites, and 30 optimization iterations. Timing includes initial observation, goal, and context encoding once, followed by all candidate evaluations in latent space.}
    \label{tab:pusht_inference_efficiency}
\end{table}

\clearpage
\subsection{Success-threshold selection and sensitivity}
\label{sec:appendix_success_thresholds}
\begin{figure}[htbp]
    \centering
    \includegraphics[width=\textwidth]{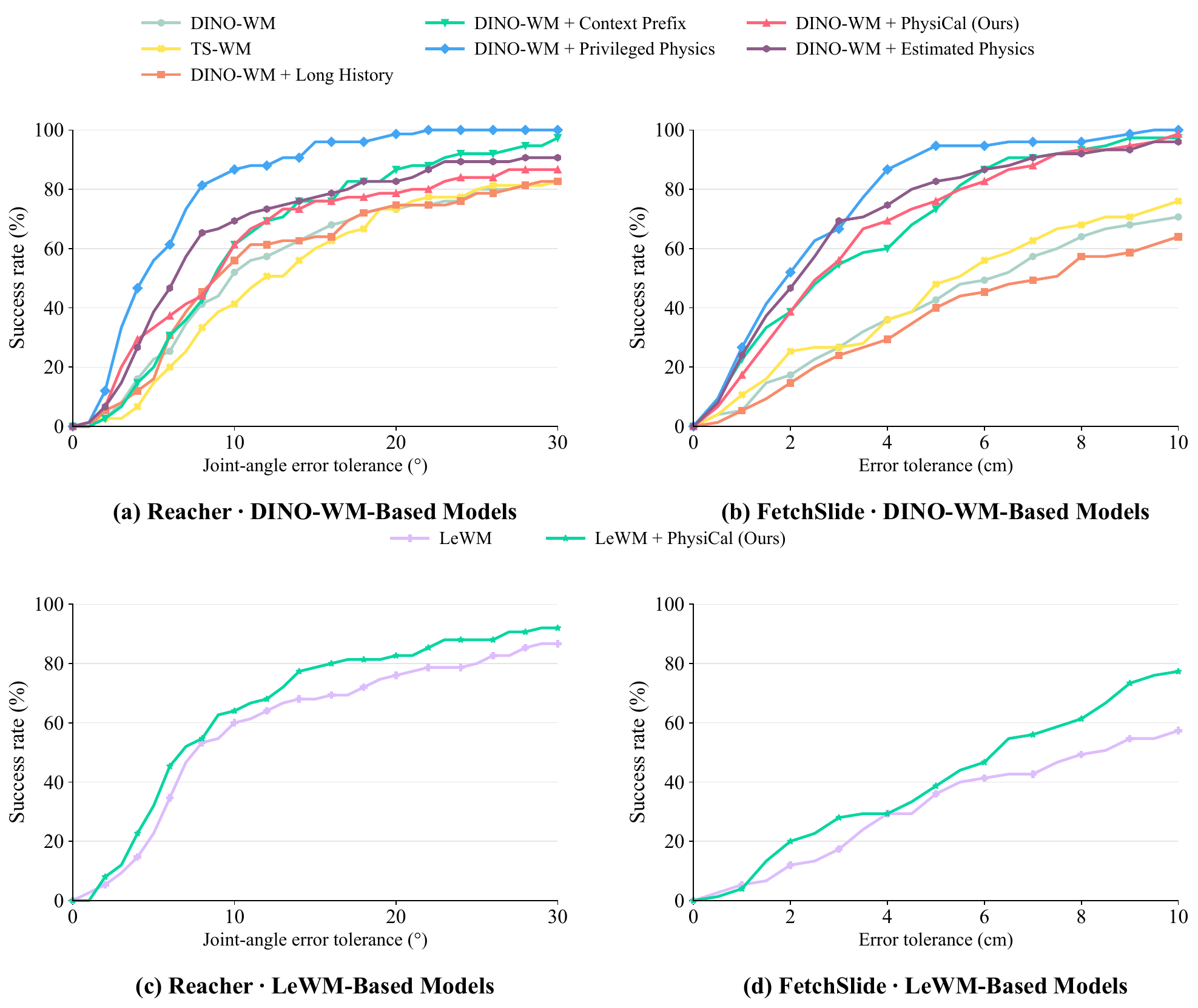}

    \caption{\textbf{Success-threshold sensitivity.} Evaluation of planning success rates across varying error tolerances for DINO-WM variants (top) and LeWM variants (bottom).
    (a) \textbf{Reacher (DINO-WM-based models)} using maximum absolute joint-angle error across joints. 
    (b) \textbf{FetchSlide (DINO-WM-based models)} using terminal puck-XY position error. 
    (c) \textbf{Reacher (LeWM-based models)} across joint-angle tolerances. 
    (d) \textbf{FetchSlide (LeWM-based models)} across puck-XY position tolerances.}
    \label{fig:success_tolerance}
\end{figure}

To better understand the planning performance in Reacher and FetchSlide (as reported in \autoref{tab:main_pusht_reacher_results} and \autoref{tab:main_beerpong_fetchslide_results}), we additionally present success rate vs. error-threshold curves in \autoref{fig:success_tolerance}.

Let the error threshold be $\tau$, measured in degrees for Reacher and centimeters for FetchSlide. 
For Reacher, success requires both joint angular position errors to satisfy $\max_{j\in\{1,2\}}|q_j^{25}-q_j^{\mathrm{goal}}|<\tau$, where $q_j^{25}$ denotes the angle of joint $j$ after executing the 25 planned controls. 
For FetchSlide, the success criterion is that terminal puck-position error $\lVert\mathbf{p}_{xy}^{\mathrm{final}}-\mathbf{p}_{xy}^{\mathrm{goal}}\rVert_2\leq\tau$. 
For the results shown in \autoref{tab:main_pusht_reacher_results} and \autoref{tab:main_beerpong_fetchslide_results}, the error tolerance $\tau$ is $10^\circ$ for Reacher and $3\,\mathrm{cm}$ for FetchSlide.

As shown in \autoref{fig:success_tolerance}, 
in both the Reacher and the FetchSlide environments, our method  improves performance over the baseline world models (DINO-WM and LeWM), as evidenced by comparing DINO-WM (deep blue) and DINO-WM + ours (green).
Our method improves substantially over the unprivileged baselines in FetchSlide. The gains are modest in the Reacher environment, likely due to its highly constrained state space, which consists of only two rotational joints.

\clearpage

\begingroup
\setlength{\intextsep}{6pt}
\raggedbottom
\section{Additional Qualitative Results}
\label{sec:appendix_qualitative_results}

In this section, we provide additional qualitative results for both prediction and planning. We first present the in-context interaction clips that are used to generate the qualitative results shown in the main paper in Appendix~\ref{sec:appendix_main_context}. 
We also append the uncropped version of \autoref{fig:5-main-pusht-fetchslide} (BeerPong qualitative results) in Appendix~\ref{sec:appendix_beerpong_uncropped}. 

In Appendix~\ref{sec:appendix_extrapolation_qualitative}, we show qualitative results of the extrapolation experiments as introduced in Section \ref{sec:experiment:analysis}.
We also provide additional qualitative open-loop rollout results in Appendix~\ref{sec:appendix_openloop_qualitative} as well as planning results in Appendix~\ref{sec:appendix_planning_qualitative}. 
For each result, we also provide the corresponding in-context interaction clips that \method and other context-based baselines use.

\subsection{Context Clips for Main-Text Examples}
\label{sec:appendix_main_context}
We present the in-context interaction clips corresponding to the main-text qualitative results of Push-T~(\autoref{fig:5-main-pusht-fetchslide}), BeerPong~(\autoref{fig:5-main-pusht-fetchslide}), and FetchSlide~(\autoref{fig:6-fetchslide-main})  as in \autoref{fig:context-main-pusht}, \autoref{fig:context-main-beerpong}, and \autoref{fig:context-main-fetchslide}, respectively. 
For each example, all context-conditioned models receive the same interaction clip as input. \method encodes this clip into a \textit{physics descriptor token}, while in the Context Prefix baseline, the same clip is directly fed into the future predictor $f_\theta$.

\begin{figure}[H]
    \centering
    \includegraphics[width=.92\linewidth]{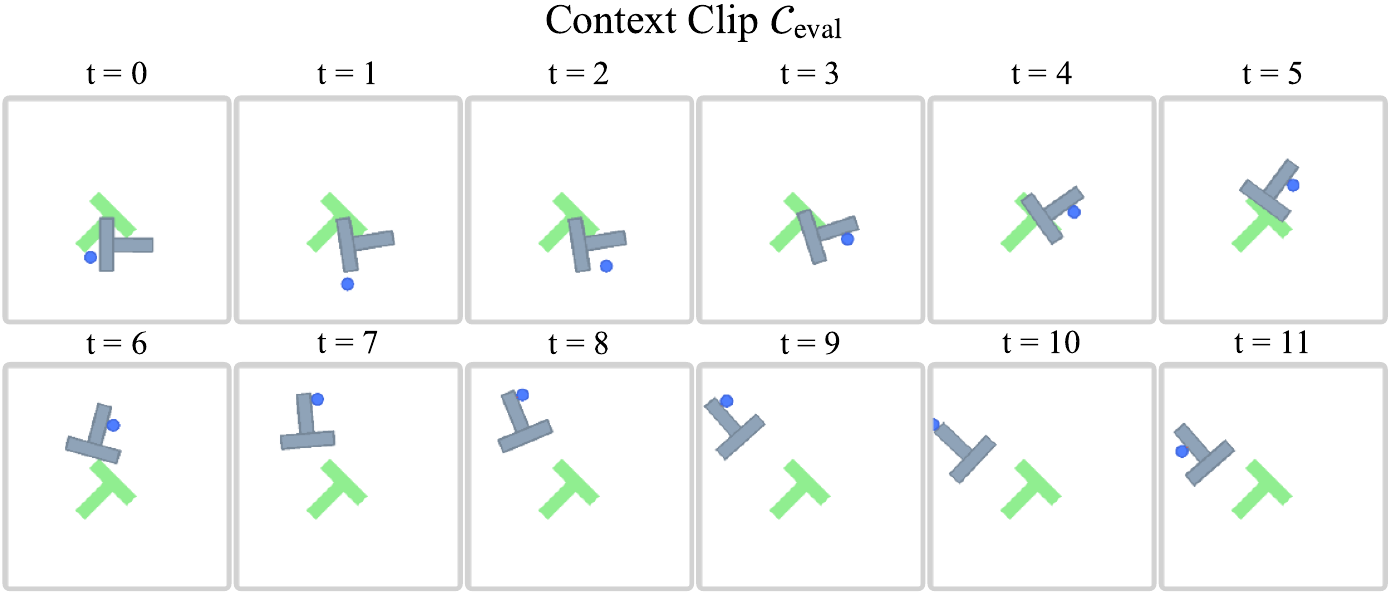}
    \vspace{-7pt}
    \caption{\textbf{Qualitative Example of an In-Context Interaction Clip $\mathcal{C}_\text{eval}$ used in the Push-T Environment.} We provide the in-context interaction clip used by +\shortmethod (Ours) to generate the main-text rollout results in \autoref{fig:5-main-pusht-fetchslide} (a). This clip is collected via the same policy (i.e., a CoM-agnostic PPO policy) that is used to generate the training dataset. This clip is encoded by the context encoder $\ctxenc$ into a \textit{physics descriptor token}, which is used to calibrate the world model's predictions.}
    \label{fig:context-main-pusht}
    \vspace{-10pt}
\end{figure}

\begin{figure}[H]
    \centering
    \includegraphics[width=.92\linewidth]{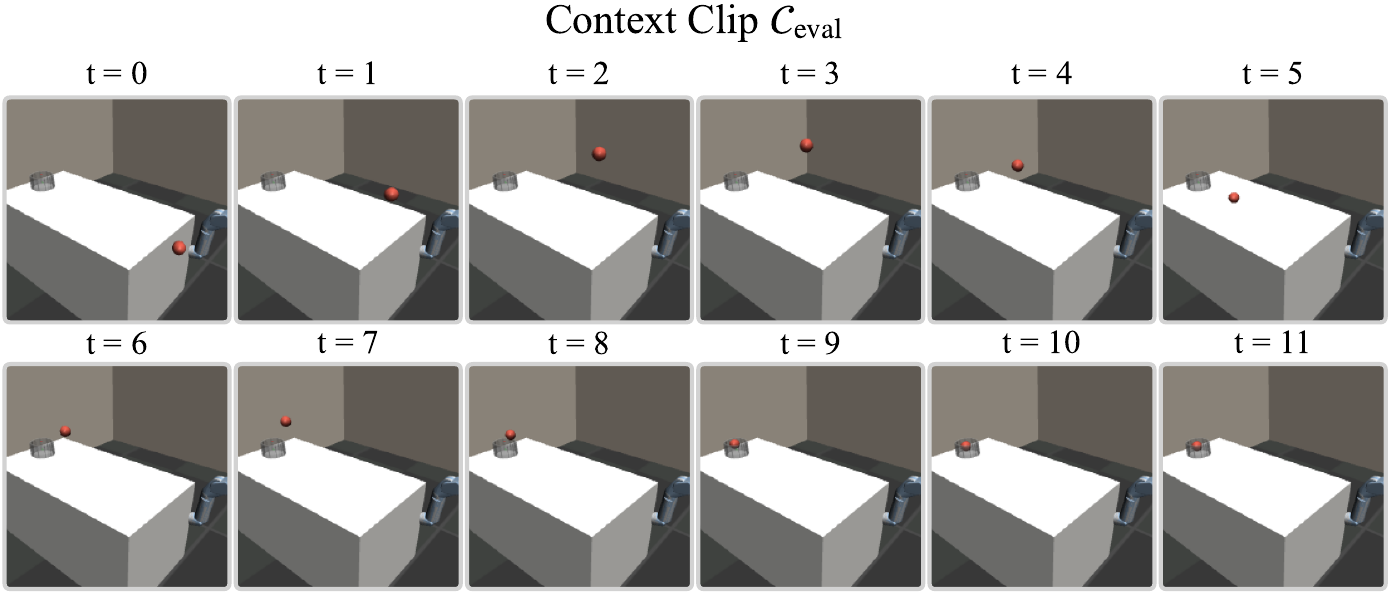}
    \vspace{-7pt}
    \caption{\textbf{Qualitative Example of an In-Context Interaction Clip $\mathcal{C}_\text{eval}$ used in the BeerPong Environment.} We show the in-context interaction clip used by +\shortmethod (Ours) to produce the main-text rollout results in \autoref{fig:5-main-pusht-fetchslide} (b). This clip is collected by a randomly sampled throwing action. 
    The context encoder $g_\phi$ subsequently takes this clip as input and encodes it into a \textit{physics descriptor token} \zp. As \zp carries the test scene's transition dynamics inferred from the interaction clip, it can effectively ground the world model's future predictions in the actual physical dynamics.}
    \label{fig:context-main-beerpong}
\end{figure}

\clearpage
\begin{figure}[H]
    \centering
    \includegraphics[width=.8\linewidth]{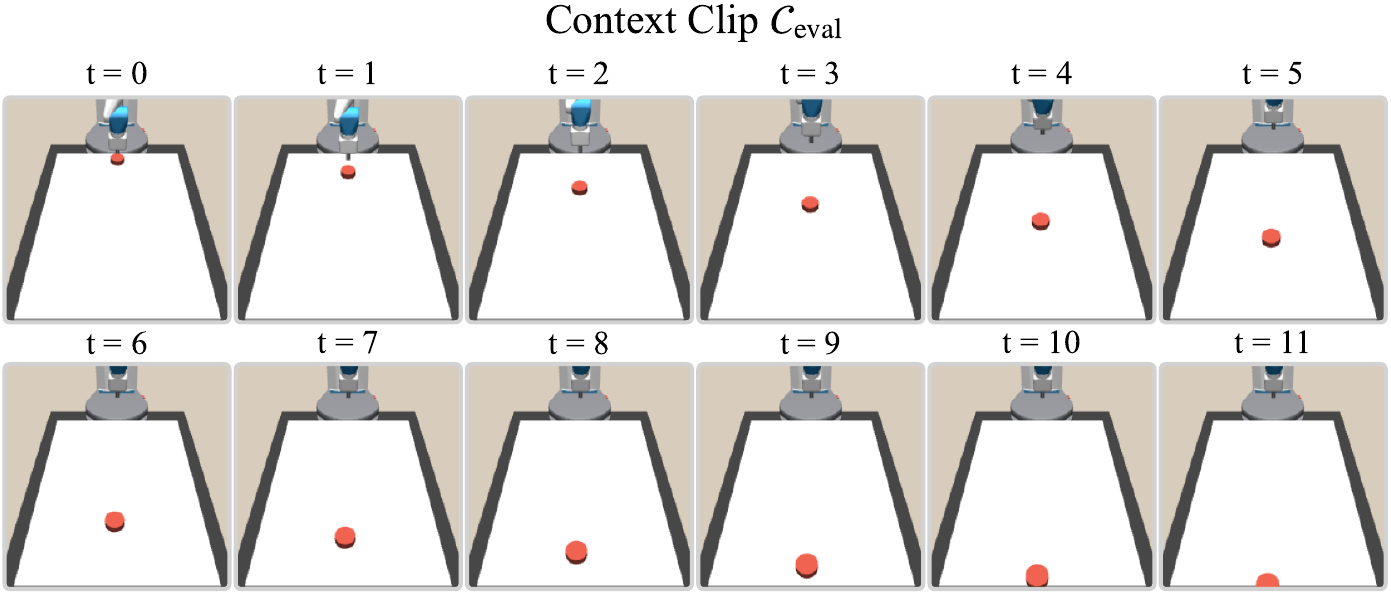}
    \caption{\textbf{Qualitative Example of an In-Context Interaction Clip $\mathcal{C}_\text{eval}$ in the FetchSlide Environment.} We present the in-context interaction clip used by \method to generate the main-text prediction results in \autoref{fig:6-fetchslide-main}. This clip is collected via a random strike action and is encoded by the context encoder $g_\phi$ into a \textit{physics descriptor token}, which is used to calibrate the world model's predictions.}
    \label{fig:context-main-fetchslide}
\end{figure}

\subsection{Uncropped BeerPong Visualization}
\label{sec:appendix_beerpong_uncropped}
In \autoref{fig:5-main-pusht-fetchslide}(b), we crop the predicted frames for a clearer view. In this section, we provide the uncropped version of the model's predictions, as shown in \autoref{fig:beerpong uncrop}.

\begin{figure}[H]
    \centering
    \includegraphics[width=.68\linewidth]{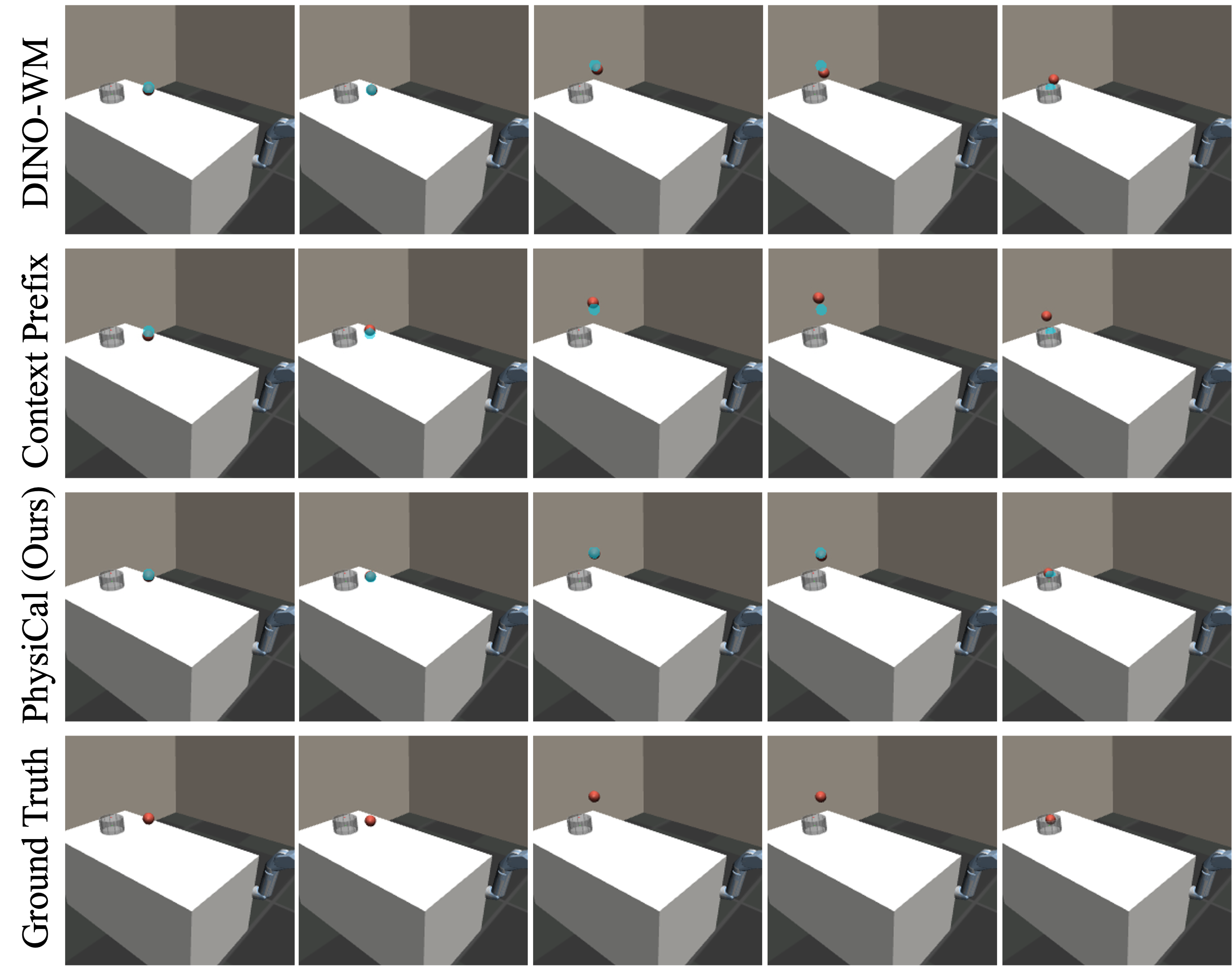}
    \caption{\textbf{Uncropped Open-Loop Rollouts in the BeerPong Environment.} We present the full-scene view of the BeerPong rollouts shown in \autoref{fig:5-main-pusht-fetchslide}(b), displaying the predicted ball trajectories relative to the table and cup. DINO-WM + Ours accurately predicts the ball's bounce, while DINO-WM and DINO-WM + Context Prefix overestimate its rebound height. The semi-transparent cyan overlay marks the ground-truth ball position.}
    \label{fig:beerpong uncrop}
\end{figure}

\subsection{Qualitative Extrapolation Results}
\label{sec:appendix_extrapolation_qualitative}
We provide qualitative world models' predictions complementing the Reacher extrapolation results shown in \autoref{fig:9-extrapolate}. Following the main-text experiment, the first joint's damping lies outside the training range, while the second joint's damping remains fixed. We present the open-loop predictions in \autoref{fig:reacher-extrapolation-qualitative} and the interaction clip used by \method in \autoref{fig:context-reacher-extrapolation}. 
We observe that the prediction quality of DINO-WM + Privileged Physics deteriorates quickly after $t = 7$, while our method performs stably throughout the open-loop rollout (from $t=0$ to $t=19$).

\begin{figure}[H]
    \centering
    \includegraphics[width=.76\linewidth]{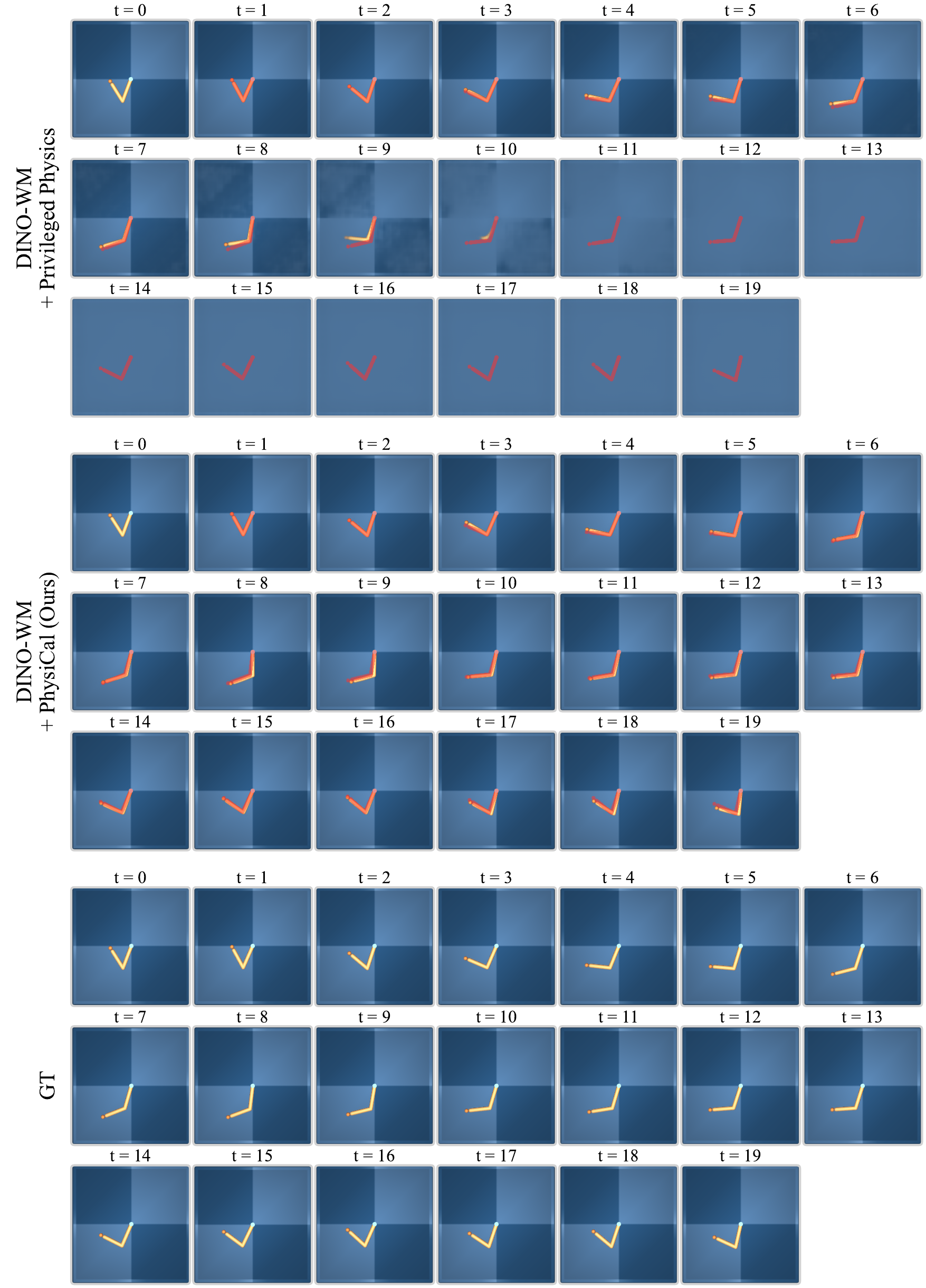}
    \vspace{-5pt}
    \caption{\textbf{Qualitative Open-Loop Rollouts in the Reacher Environment under Out-of-Distribution Physics Configurations.} We present an extrapolation example corresponding to the quantitative results in \autoref{fig:9-extrapolate}, with the first joint's damping outside the training range and the second joint's damping held fixed. Both models predict future observations autoregressively without additional ground-truth observations (i.e., feedback from the environment). DINO-WM + Privileged Physics predicts accurately from \(t=0\) to \(t=6\), but background blurring emerges at \(t=7\) and progressively intensifies through \(t=10\). From \(t=11\) to \(t=19\), its predictions collapse into a nearly uniform blue field, with no discernible arm or scene structure. In contrast, DINO-WM + Ours preserves the scene structure and continues to predict the arm's motion throughout the rollout. The corresponding interaction clip is shown in \autoref{fig:context-reacher-extrapolation}.}
    \label{fig:reacher-extrapolation-qualitative}
    \vspace{-10pt}
\end{figure}

\begin{figure}[H]
    \centering
    \includegraphics[width=.61\linewidth]{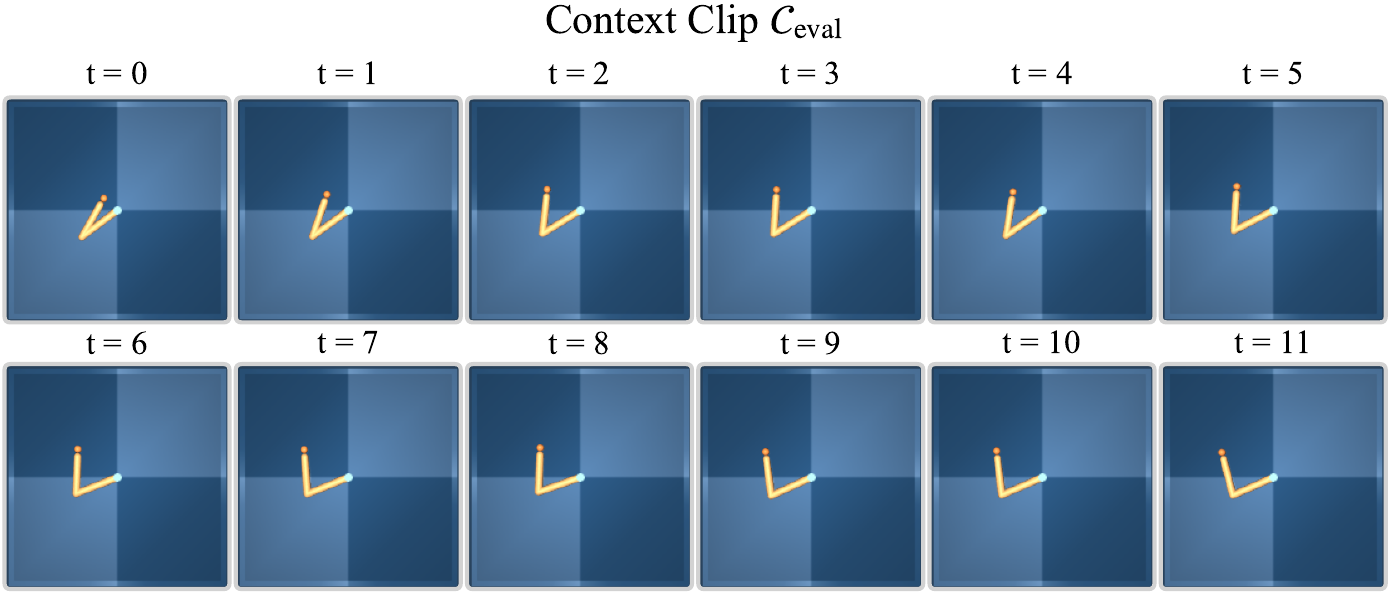}
    \vspace{-8pt}
    \caption{\textbf{In-Context Interaction Clip $\mathcal{C}_\text{eval}$ used in \autoref{fig:reacher-extrapolation-qualitative}.}
    This clip is collected using randomly sampled actions (the action space is the torques applied at the two joints of the robot arm).}
    \label{fig:context-reacher-extrapolation}
\end{figure}

\clearpage
\subsection{Additional Qualitative Open-Loop Rollout Results}
\vspace{-2pt}
\label{sec:appendix_openloop_qualitative}
We provide additional qualitative open-loop rollout results in all four environments: Push-T (\autoref{fig:pusht-openloop}), Reacher~(\autoref{fig:reacher-openloop}), BeerPong~(\autoref{fig:beerpong-openloop}), and FetchSlide~(\autoref{fig:fetchslide-openloop}).
Starting from the initial observations, each model predicts future observations autoregressively under a given sequence of evaluation actions. That is, the predicted observation will be fed back to the model as the current observation to synthesize next observation.
Each rollout result is followed by the interaction clip used by the context-conditioned models.

\begin{figure}[H]
    \vspace{-2pt}
    \centering
    \includegraphics[width=.95\linewidth]{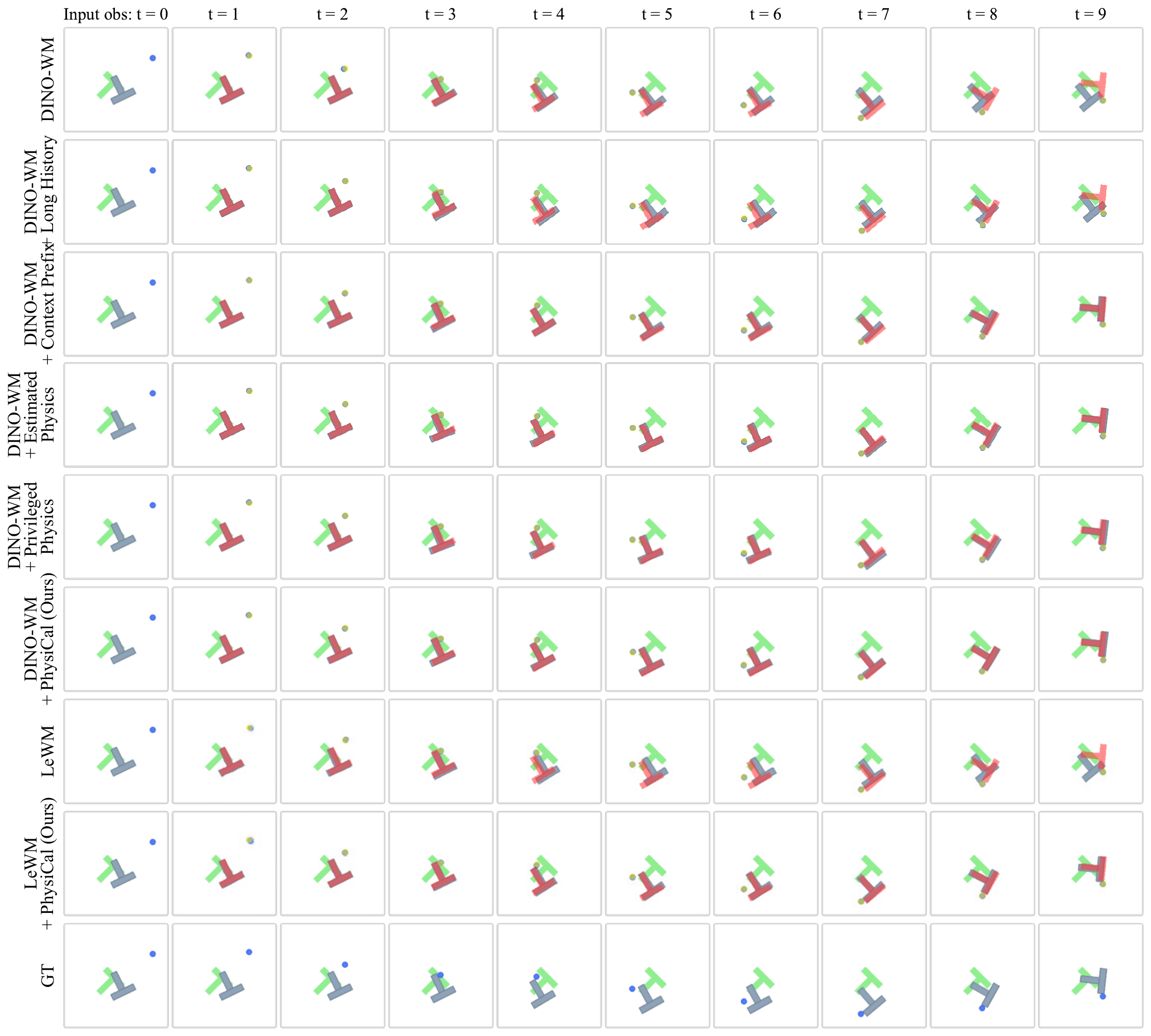}
    \vspace{-7pt}
    \caption{\textbf{Qualitative Open-Loop Rollouts in the Push-T Environment.} We present open-loop rollouts, where models are only given the initial observation and autoregressively predict future observations. All context-conditioned models use the interaction clip shown in \autoref{fig:context-pusht-openloop}. DINO-WM + Ours, LeWM + Ours, DINO-WM + Privileged Physics, and DINO-WM + Context Prefix closely match the ground-truth motion, while the remaining baselines exhibit substantial prediction errors. Though Context Prefix predicts accurately, it directly attends to the full context sequence, making its inference substantially slower.}
    \label{fig:pusht-openloop}
\end{figure}

\begin{figure}[H]
    \vspace{-9pt}
    \centering
    \includegraphics[width=.62\linewidth]{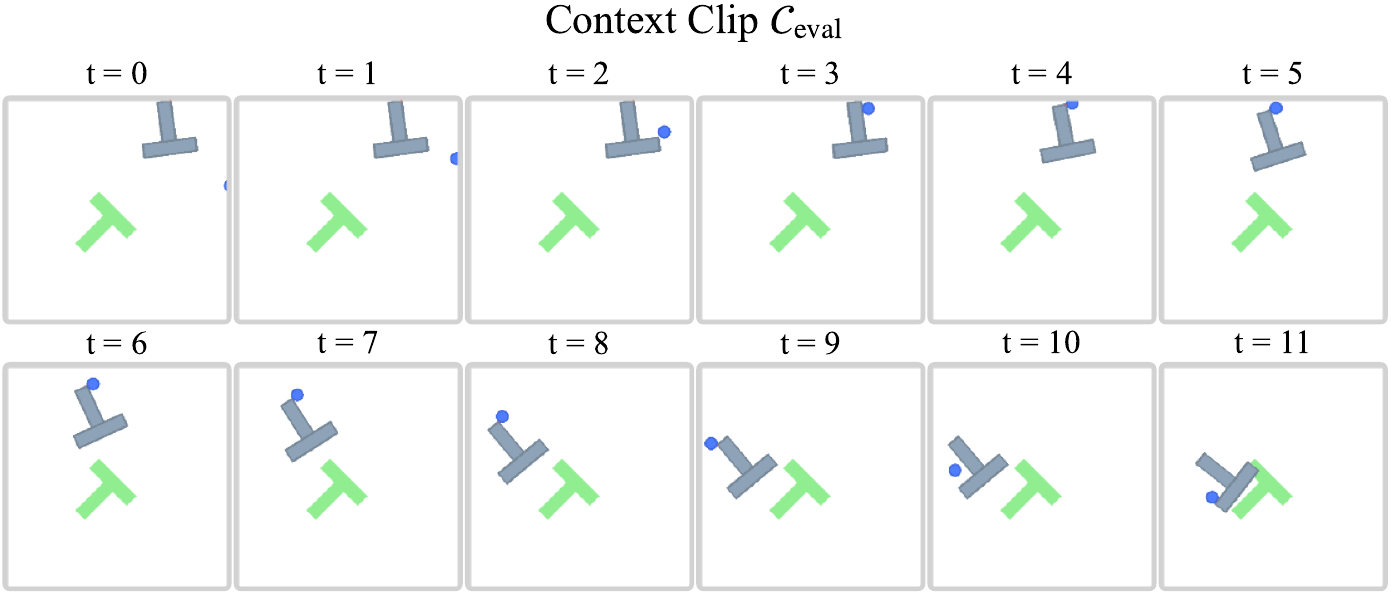}
    \vspace{-6pt}
    \caption{\textbf{Qualitative Example of an Exploratory In-Context Interaction Clip $\mathcal{C}_\text{eval}$ in the Push-T Environment.} This in-context interaction clip is used by \method to generate the open-loop rollout results in \autoref{fig:pusht-openloop}. The clip is collected using a CoM-agnostic PPO policy and encoded by the context encoder $g_\phi$ into a \textit{physics descriptor token} to calibrate the world model's predictions.}
    \label{fig:context-pusht-openloop}
\end{figure}

\begin{figure}[H]
    \centering
    \includegraphics[width=1\linewidth]{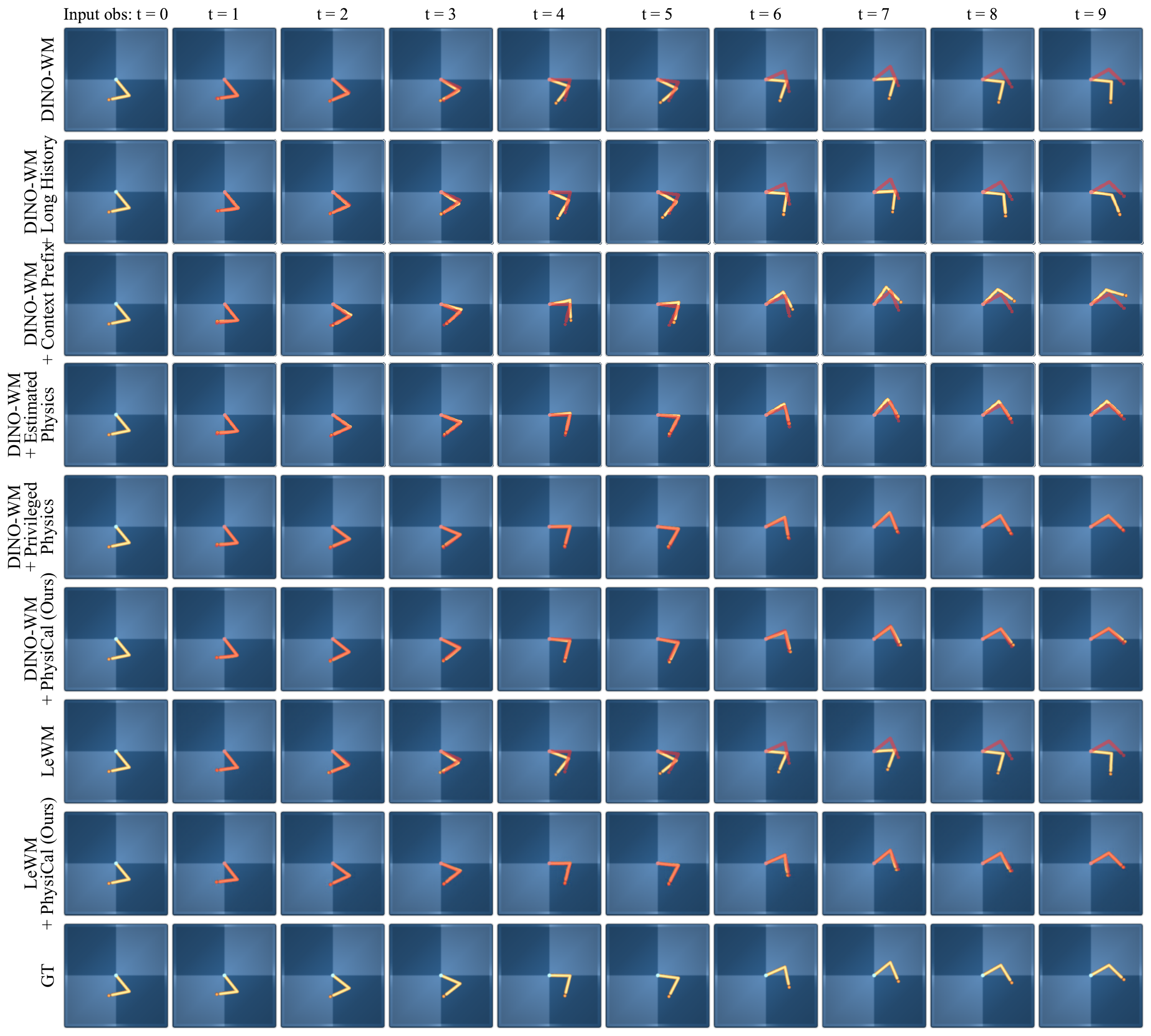}
    \vspace{-14pt}
    \caption{\textbf{Qualitative Open-Loop Rollouts in the Reacher Environment.} We show autoregressive predictions of the arm's motion without providing further ground-truth observations during the rollout. 
    All the models that ingest a context clip use the same one as presented in \autoref{fig:context-reacher-openloop}.
    DINO-WM + Ours, LeWM + Ours, and DINO-WM + Privileged Physics accurately predict the arm's motion. Context Prefix overestimates the motion, whereas the remaining baselines underestimate it, suggesting that it uses the context to adjust its predictions but remains less accurate than our models.}
    \label{fig:reacher-openloop}
\end{figure}

\begin{figure}[H]
    \vspace{-7pt}
    \centering
    \includegraphics[width=.7\linewidth]{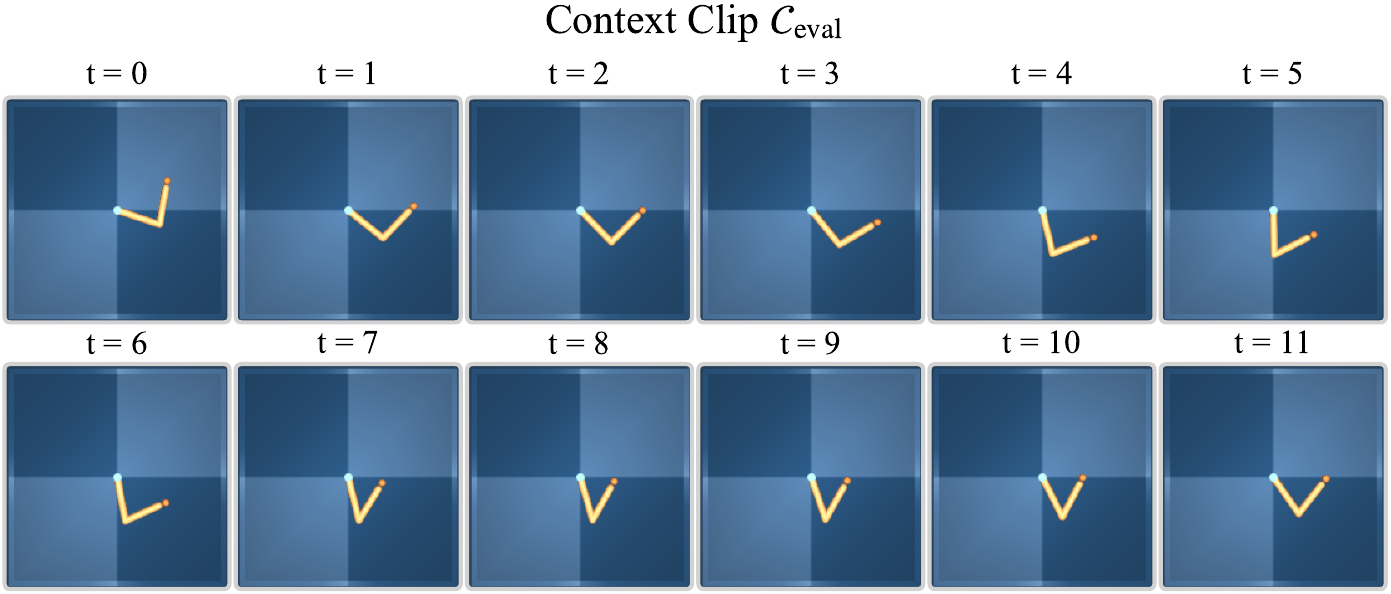}
    \vspace{-5pt}
    \caption{\textbf{Qualitative Example of an In-Context Interaction Clip $\mathcal{C}_\text{eval}$ in the Reacher Environment.} We provide the in-context interaction clip used by \method to produce the open-loop rollout results in \autoref{fig:reacher-openloop}. Generated by randomly sampled joint-torque commands, this clip is encoded by the context encoder $g_\phi$ into a \textit{physics descriptor token}, which is used to calibrate the world model's predictions.}
    \label{fig:context-reacher-openloop}
\end{figure}

\clearpage

\begin{figure}[H]
    \centering
    \includegraphics[width=1\linewidth]{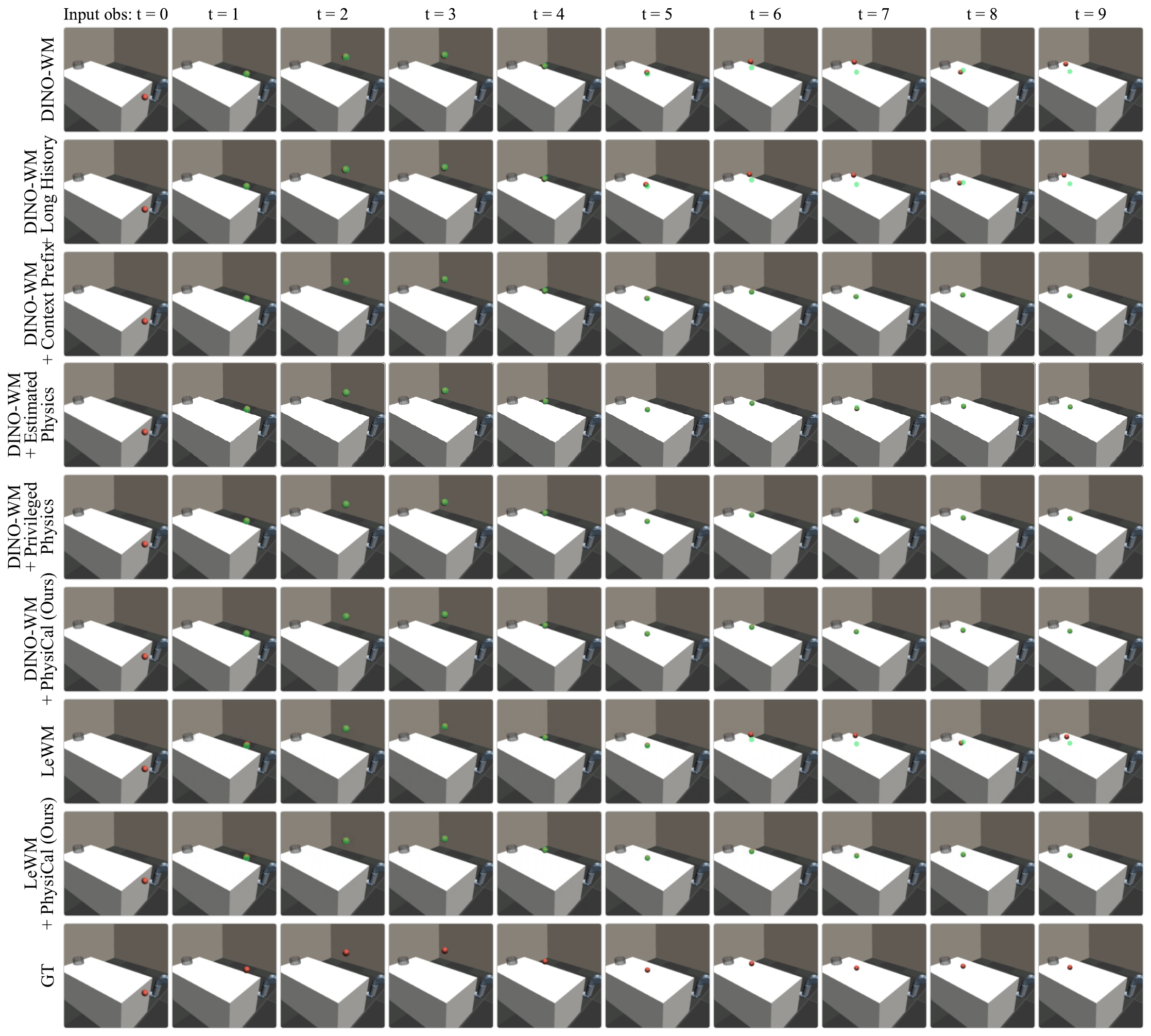}
    \caption{\textbf{Qualitative Open-Loop Rollouts in the BeerPong Environment.} We provide open-loop predictions of the ball's trajectory, generated autoregressively without subsequent ground-truth observations. All models that use context receive the interaction clip in \autoref{fig:context-beerpong-openloop}. DINO-WM + Ours, LeWM$^{*}$ + Ours, DINO-WM + Privileged Physics, and DINO-WM + Context Prefix accurately reproduce the ball's bounce, while the remaining baselines overestimate its rebound. The accuracy of Context Prefix, however, comes at the cost of substantially slower inference due to direct attention over the full context sequence. LeWM$^{*}$ denotes TC-LeWM in BeerPong, as we find that this variant can significantly boost the performance of LeWM in BeerPong.}
    \label{fig:beerpong-openloop}
\end{figure}

\begin{figure}[H]
    \centering
    \includegraphics[width=.7 \linewidth]{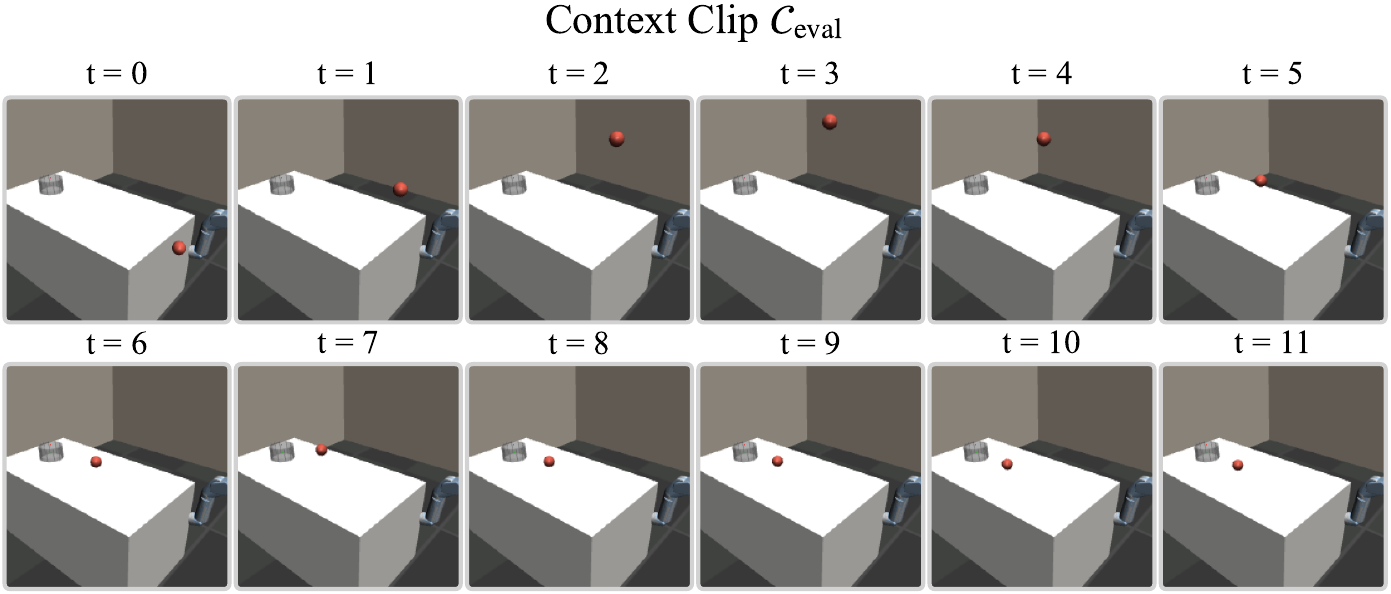}
    \caption{\textbf{Qualitative Example of an Exploratory In-Context Interaction Clip $\mathcal{C}_\text{eval}$ in the BeerPong Environment.} We present the in-context interaction clip used by \method to generate the open-loop rollout results in \autoref{fig:beerpong-openloop}. The context encoder $g_\phi$ encodes this clip, collected using a randomly sampled throw action, into a \textit{physics descriptor token} that calibrates the world model's predictions.}
    \label{fig:context-beerpong-openloop}
\end{figure}

\clearpage
\begin{figure}[H]
    \centering
    \includegraphics[width=.99\linewidth]{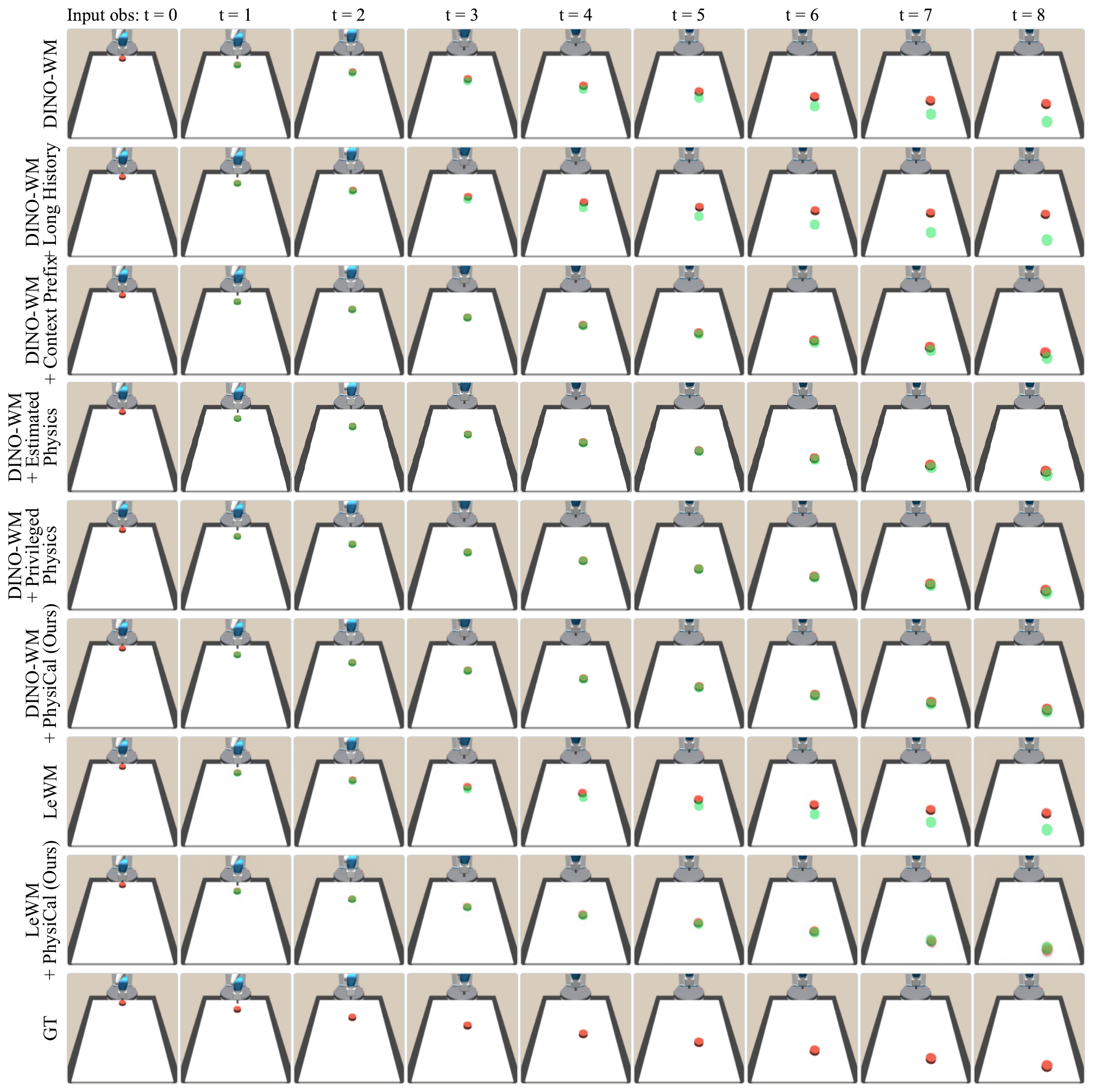}
    \vspace{-7pt}
    \caption{\textbf{Qualitative Open-Loop Rollouts in the FetchSlide Environment.} We present an eight-frame open-loop rollout, in which each model predicts future observations autoregressively without receiving additional ground-truth observations. The interaction clip shown in \autoref{fig:context-fetchslide-openloop} is shared by all context-conditioned models. DINO-WM + Ours, LeWM + Ours, and DINO-WM + Privileged Physics closely match the ground-truth puck motion. DINO-WM + Context Prefix slightly underestimates its movement, while the remaining baselines without context exhibit more pronounced underestimation.}
    \label{fig:fetchslide-openloop}
\end{figure}

\begin{figure}[H]
    \centering
    \includegraphics[width=.65\linewidth]{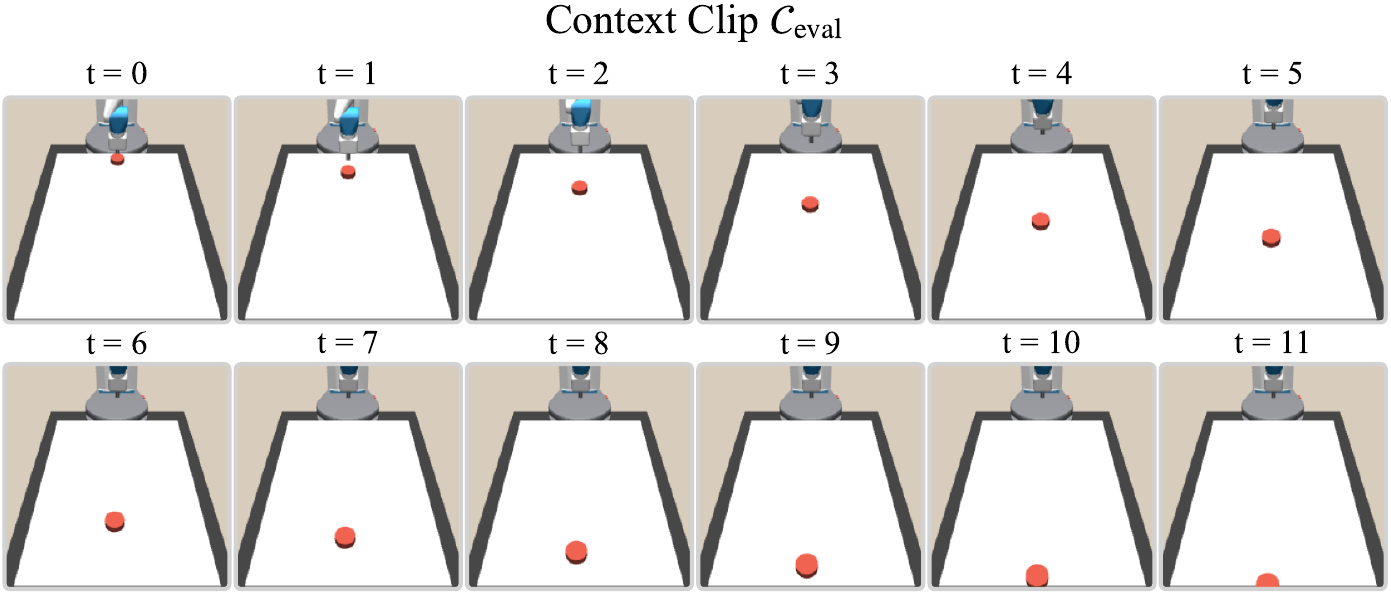}
    \vspace{-7pt}
    \caption{\textbf{Qualitative Example of an Exploratory In-Context Interaction Clip $\mathcal{C}_\text{eval}$ in the FetchSlide Environment.} We provide the in-context interaction clip used by the context-conditioned methods to generate the results in \autoref{fig:fetchslide-openloop}. This clip is collected using a randomly sampled strike action and subsequently encoded by the context encoder $g_\phi$ into a \textit{physics descriptor token}, which is used to calibrate the world model's predictions.}
    \label{fig:context-fetchslide-openloop}
\end{figure}

\clearpage
\subsection{Additional Planning Results}
\label{sec:appendix_planning_qualitative}
\vspace{-2pt}
We present additional planning comparisons of both DINO-WM- and LeWM-based models. For each example, we 
use the world model with CEM to optimize a sequence of actions that reaches the goal observation. Planned actions are executed in the simulated environment. We present both the world models' predicted observations (top row) and the actual execution in the simulated environments (bottom row). Each planning example is followed by the corresponding interaction clip.

\begin{figure}[H]
    \centering
    \vspace{-3pt}
    \includegraphics[width=.78\linewidth]{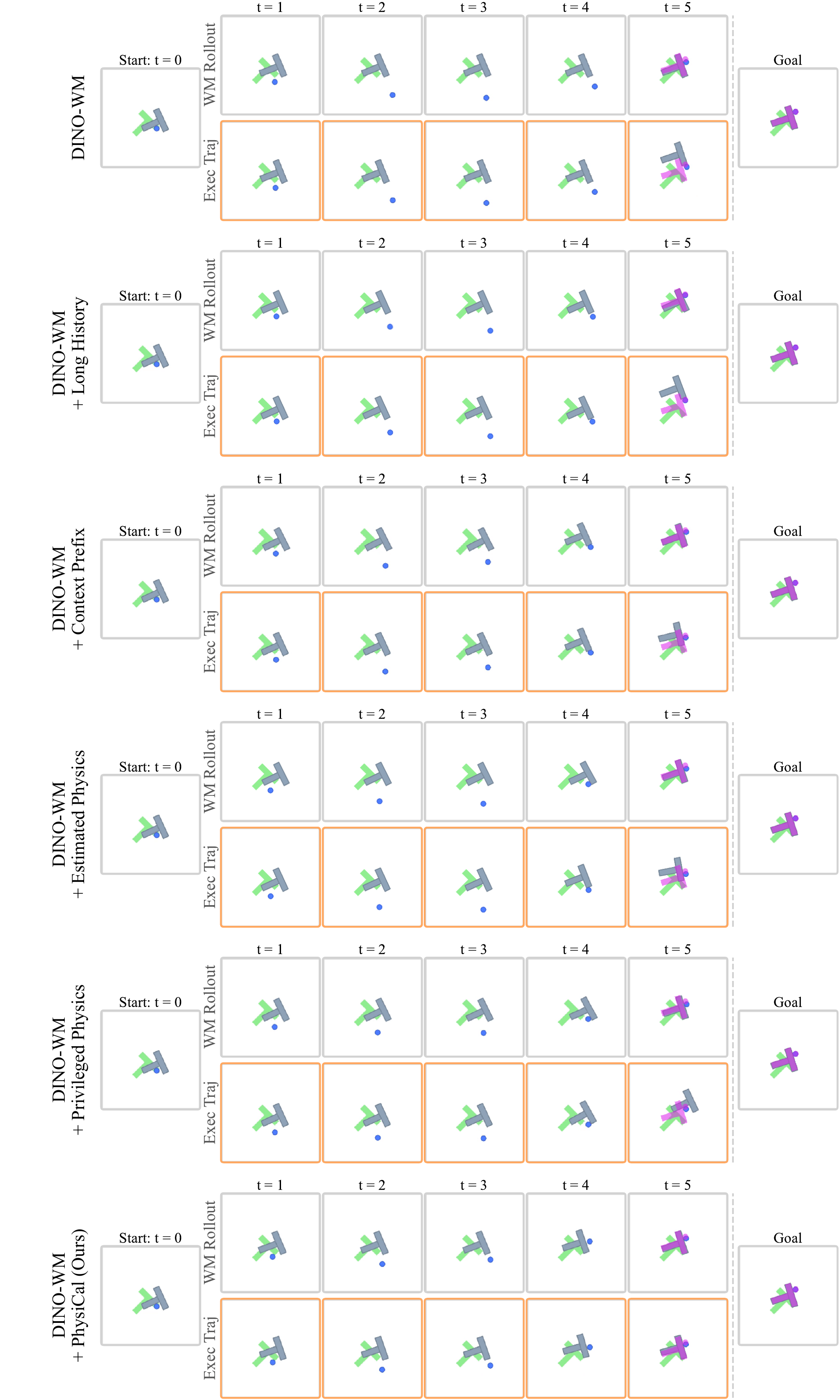}
    \vspace{-6pt}
    \caption{\textbf{Qualitative Planning Rollouts in the Push-T Environment, DINO-WM-Based Models.} We use CEM to optimize action sequences toward a goal observation and execute them in the environment, comparing model predictions with the resulting motion. All context-conditioned models use the interaction clip shown in \autoref{fig:context-pusht-dino-plan}. DINO-WM + Ours closely matches the executed motion, while the remaining models misrepresent the T-block's translational and rotational dynamics.}
    \label{fig:pusht-dino-plan}
\end{figure}

\clearpage
\begin{figure}[H]
    \centering
    \includegraphics[width=.69\linewidth]{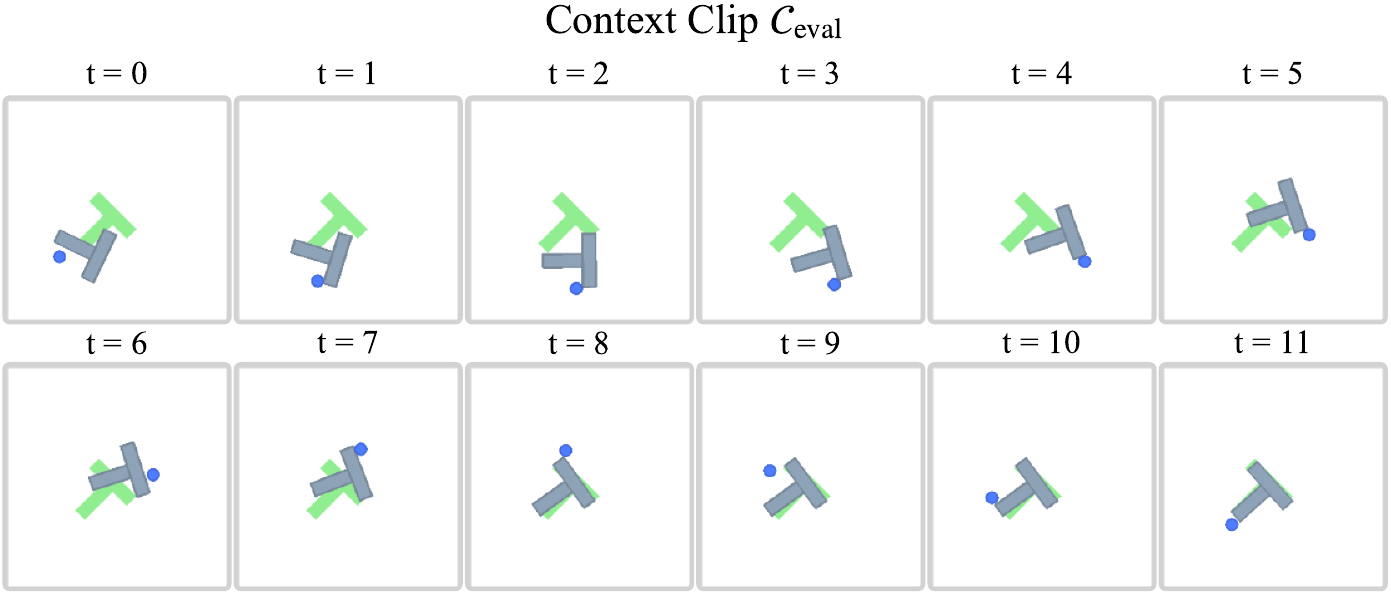}
    \vspace{-7pt}
    \caption{\textbf{Qualitative Example of an In-Context Interaction Clip $\mathcal{C}_\text{eval}$ in the Push-T Environment.} We provide the in-context interaction clip used by \method to generate the planning results in \autoref{fig:pusht-dino-plan}. Collected using a CoM-agnostic PPO policy, this clip is encoded by the context encoder $g_\phi$ into a \textit{physics descriptor token}, which is used to calibrate the world model's predictions during planning.}
    \label{fig:context-pusht-dino-plan}
\end{figure}

\begin{figure}[H]
    \centering
    \includegraphics[width=.95\linewidth]{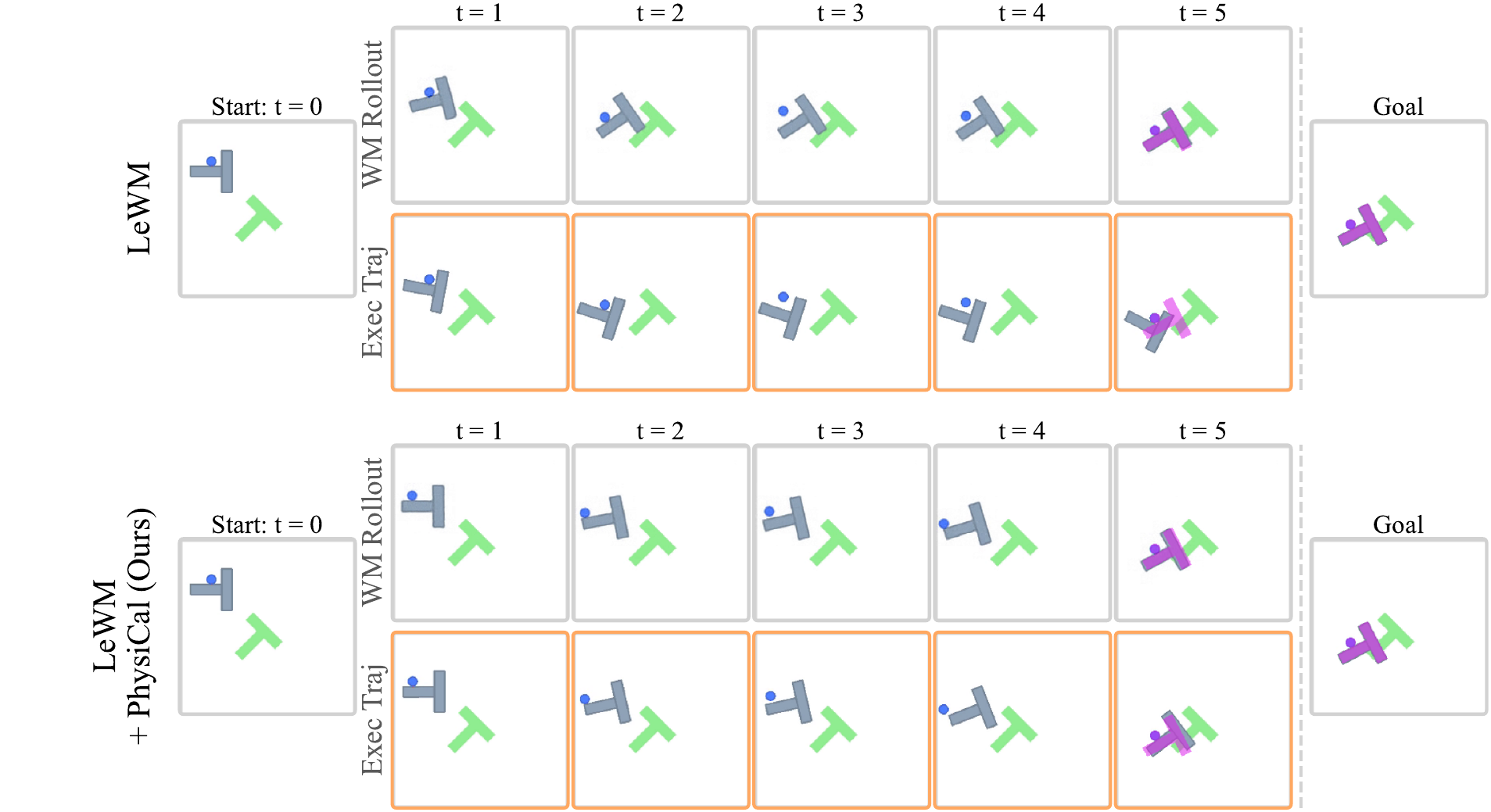}
    \caption{\textbf{Qualitative Planning Rollouts in the Push-T Environment, LeWM-Based Models.} Action sequences generated by CEM to reach a goal observation are executed in the environment and compared with the world models' predictions. LeWM + Ours uses the interaction clip presented in \autoref{fig:context-pusht-le-plan} and accurately predicts the resulting motion. In contrast, LeWM underestimates the effect of the push at \(t=3\), with the T-block moving and rotating substantially farther than predicted at \(t=4\).}
    \label{fig:pusht-le-plan}
\end{figure}

\begin{figure}[H]
    \centering
    \includegraphics[width=.69\linewidth]{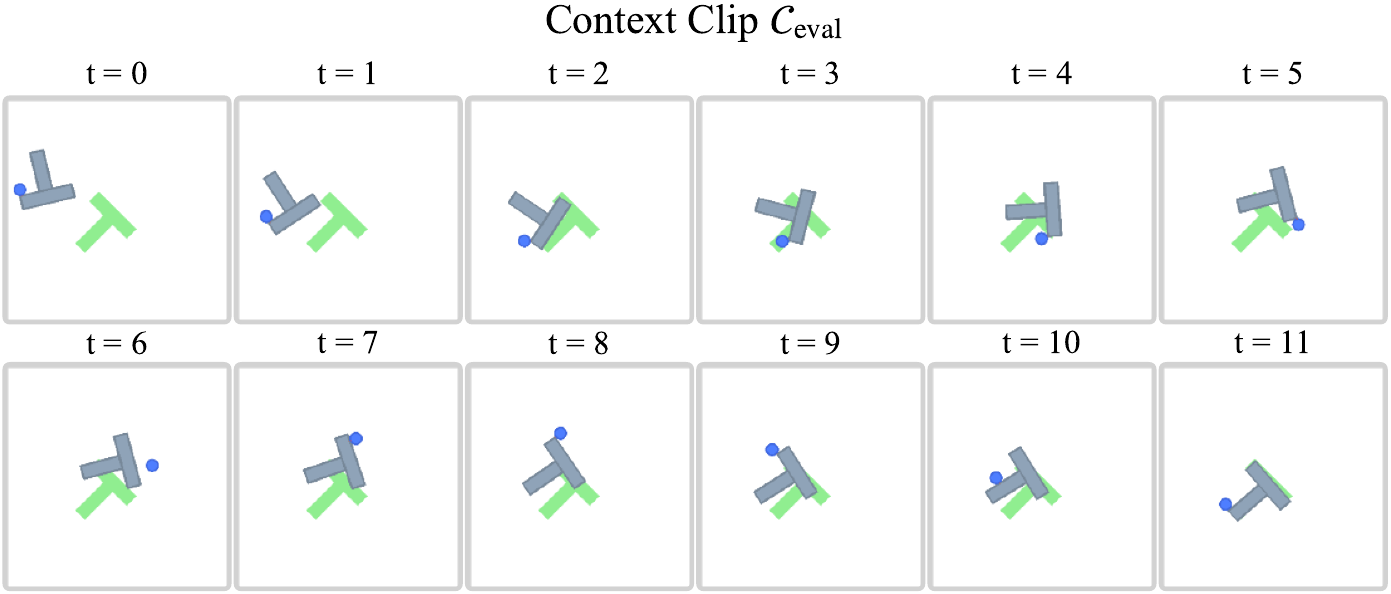}
    \vspace{-5pt}
    \caption{\textbf{Qualitative Example of an Exploratory In-Context Interaction Clip $\mathcal{C}_\text{eval}$ in the Push-T Environment.} We show the in-context interaction clip used by \method to produce the planning results in \autoref{fig:pusht-le-plan}. This clip is collected using a CoM-agnostic PPO policy. The proposed method will use its context encoder $g_\phi$ to encode this clip into a \textit{physics descriptor token}, which is used to calibrate the world model's predictions during planning.}
    \label{fig:context-pusht-le-plan}
\end{figure}

%% file: sections/pseudo_code.tex
\vspace{-8pt}
\subsection{Training and Inference Pseudocode of \method}
\label{sec:pseudo_code}

We present the pseudocode for training \method in Algorithm~\ref{alg:icp_training}, which is also described in Section~\ref{sec:method:training_without_physics_labels}.
Training \method requires neither ground-truth physics labels nor an analytical dynamics model; the only supervision structure it consumes is the group partition of the dataset (where data collected within environments sharing identical physics are grouped together). 

Hence, unlike Estimated Physics$^\dagger$ or 
Privileged Physics$^\ddagger$, \method learns to calibrate directly from the dynamics patterns in the data: it does not need to know which physics properties actually govern the transition dynamics nor their corresponding values.
Because every episode in $\mathcal{G}_j$ shares the same underlying transition dynamics $\mathcal{T}_{\xi_j}$, if both the context prompt and the train target segment are sampled from the same group, context encoder $\ctxenc$ can minimize the prediction loss by encoding context-specific physical dynamics $\xi_j$ from the prompt into \zp.
Note that we draw the interaction prompt from an episode distinct from the trajectory segment used for future prediction so as to prevent the prompt from carrying information about the future.

During training, the predictor is teacher-forced on the visual encoder representations under causal masking (e.g., $z_0$ cannot attend to $z_1$), with prediction loss gradients flowing jointly into both $\theta$ and $\phi$. The visual decoder $D_\text{v}$ is trained \textit{separately} using an $\ell_2$ reconstruction loss to map predicted latents $\hat{z}_t$ back to pixel-space images.

\begin{algorithm}[H]
\caption{Training \method}
\label{alg:icp_training}
\setstretch{1.0}
\begin{algorithmic}[1]
  \STATE \textbf{Require:} train dataset $\mathcal{D}_{\text{train}} = \{\mathcal{G}_j\}_{j=1}^M$ of $M$ physics configurations, context encoder $\ctxenc$, visual encoder $E_\text{v}$, action encoder $E_\text{a}$, dynamics predictor $f_\theta$, input window length $K$, prompt length $T_c$, training steps $N$
  \For{ $i = 1 \to N$ }
    \STATE $\mathcal{G}_j \sim \mathcal{D}_{\text{train}}$, sample a data group from dataset $\mathcal{D}_\text{train}$
    \STATE $(o_{0:K}, a_{0:K-1}) \sim \mathcal{G}_j$, sample a trajectory segment of length $K+1$ for training  
    \STATE \commentcode{Sample interaction prompt from distinct episode in $\mathcal{G}_j$}
    \STATE $\mathcal{C}_{\text{tr}}^j = \{(o_k^c, a_k^c)\}_{k=1}^{T_c} \sim \mathcal{G}_j$, with $\text{episode\_id}(\mathcal{C}_{\text{tr}}^j) \neq \text{episode\_id}(o_{0:K})$
    \STATE $\zp \leftarrow \ctxenc(\mathcal{C}_{\text{tr}}^j)$ \commentcode{encode physics descriptor token}
    \STATE $z_{0:K} \leftarrow E_\text{v}(o_{0:K})$ \commentcode{encode visual observation tokens}
    \STATE $e_{0:K-1}^a \leftarrow E_\text{a}(a_{0:K-1})$ \commentcode{encode action embeddings}
    \STATE \commentcode{Predict future latents $\hat{z}_{1:K}$ with causal masks (e.g., $z_0$ cannot attend to $z_1$) }
    \STATE $\hat{z}_{1:K} \leftarrow f_\theta\left(\{(z_t, e_t^a)\}_{t=0}^{K-1},\; \zp\right)$
    \STATE $\mathcal{L} \leftarrow \frac{1}{K}\sum_{t=1}^K \|\hat{z}_t - z_t\|_2^2$
    \STATE Backprop to update $(\theta, \phi)$ and $E_\text{a}$ using $\mathcal{L}$ 
  \EndFor
  \STATE \textbf{return} \ctxenc, $E_\text{v}, E_\text{a}, f_\theta$
\end{algorithmic}
\end{algorithm}
\vspace{-5pt}

We outline the procedure for open-loop rollout inference with \method in Algorithm~\ref{alg:icp_inference}, as detailed in Section~\ref{sec:method:inference}. At test time, the descriptor is extracted once prior to the rollout loop and subsequently held fixed. Inside the loop, autoregressive rollout runs similarly to the base world model except for the addition of the conditioning physics descriptor token, preserving per-step latency at the level of the unconditioned base predictor (Appendix~\ref{sec:appendix_inference_time}).
Latent-to-pixel decoding through $D_\text{v}$ is only required for visual evaluation; our visual planning directly operates in the latent space, $\hat{z}$.

\noindent
\begin{algorithm}[H]
\caption{Open-loop Rollout with \method}
\label{alg:icp_inference}
\setstretch{1.0}
\begin{algorithmic}[1]
  \STATE \textbf{Models:} trained context encoder $\ctxenc$, visual encoder $E_\text{v}$, action encoder $E_\text{a}$, dynamics predictor $f_\theta$, visual decoder $D_\text{v}$
  \STATE \textbf{Input:} interaction prompt $\mathcal{C}_{\text{eval}} = \{(o_k^e, a_k^e)\}_{k=1}^{T_c}$, initial observation $o_{0}$, input actions $a_{0:T-1}$, input window length $K$, open-loop rollout horizon $T$
  \STATE $\zp^\text{ev} \leftarrow \ctxenc(\mathcal{C}_{\text{eval}})$ \commentcode{extract the physics descriptor token of the test scene}
  \STATE \commentcode{encode the initial observation and set $\hat{z}_0 = z_0$ so that line~\ref{line:icp_rollout_predict} is well defined at $t=0$}
  \STATE $z_0 \leftarrow E_\text{v}(o_0)$ ~;~ $\hat{z}_0 \leftarrow z_0$
  \For{ $t = 0 \to T-1$ }
    \STATE $e_t^a \leftarrow E_\text{a}(a_t)$ \commentcode{encode the $t$-th action embedding}
    \STATE $\hat{z}_{t+1} \leftarrow f_\theta\left( \{(\hat{z}_i, e_i^a)\}_{i=\max(0,\, t - K + 1)}^t,\; \zp^\text{ev} \right)$ \commentcode{predict future latents autoregressively} \label{line:icp_rollout_predict}
    \STATE $\hat{o}_{t+1} \leftarrow D_\text{v}(\hat{z}_{t+1})$ \commentcode{decode the predicted latent to image}
  \EndFor
  \STATE \textbf{return} predicted future latents $\hat{z}_{1:T}$ and images $\hat{o}_{1:T}$
\end{algorithmic}
\end{algorithm}
\vspace{-5pt}

%% file: sections/cem_pseudo_code.tex
\clearpage
\subsection{Visual Latent Planning with Cross-Entropy Method (CEM)}
\label{sec:appendix_cem}
Following DINO-WM~\citep{zhou2025dino}, we employ the Cross-Entropy Method (CEM) for zero-shot visual latent planning (see Algorithm~\ref{alg:icp_cem} for pseudocode).
Given the current observation $o_0$ and the goal observation $o_g$, both represented as RGB images, we first encode them into latent space via the visual encoder $E_\text{v}$, namely $z_0$ and $z_g$.
We first collect an interaction clip $\mathcal{C}_{\text{eval}}$ as the context, either through 
purely random actions or a task-agnostic policy for ensuring sufficient contacts. 
We then extract the physics descriptor token of the test scene $\zp^\text{ev} \leftarrow \ctxenc(\mathcal{C}_{\text{eval}})$.
As within each planning task, the physics configuration of the scene is constant, we can use the same $\zp^\text{ev}$ across all the CEM iterations of one task, reducing the planning time. 

At each CEM iteration, we sample $N$ (e.g., $N=300$) candidate action sequences of horizon $T$ from a Gaussian distribution that is independent across time steps and action dimensions, parameterized by a mean $\boldsymbol{\mu} \in \mathbb{R}^{T \times d_a}$ and a standard deviation $\boldsymbol{\sigma} \in \mathbb{R}^{T \times d_a}$ (initialized to $\boldsymbol{\mu}=\mathbf{0}, \boldsymbol{\sigma}=\mathbf{1}$), where $d_a$ is the action dimension. 
For each candidate sequence, we unroll the physics-calibrated world model $f_\theta$ autoregressively in the latent space and evaluate the corresponding cost $C$, which we define as the mean squared error (MSE) between the final predicted latent $\hat{z}_T$ and the goal latent $z_g$, as in Line \ref{app:algo:cem_compute_cost}.
We then select the top $M_\text{elite}$ (e.g., $M_\text{elite}=30$) action sequences with the lowest costs and refit $\boldsymbol{\mu}$ and $\boldsymbol{\sigma}$ to their element-wise mean and standard deviation.
After $J$ optimization iterations (e.g., $J=30$), the planner outputs the updated elite-mean action sequence $\boldsymbol{\mu} = a_{0:T-1}^*$. 
We then execute $a_{0:T-1}^*$ in the environment. 
See \autoref{tab:cem_params} for the CEM hyperparameters in each environment.

\vspace{-9pt}
\noindent
\begin{algorithm}[H]
\caption{Visual Latent Planning with \method via CEM}
\label{alg:icp_cem}
\setstretch{1.0}
\begin{algorithmic}[1]
  \STATE \textbf{Models:} context encoder $\ctxenc$, visual encoder $E_\text{v}$, action encoder $E_\text{a}$, dynamics predictor $f_\theta$
  \STATE \textbf{Input:} interaction prompt $\mathcal{C}_{\text{eval}} = \{(o_k^e, a_k^e)\}_{k=1}^{T_c}$, current / goal observation $o_0$ / $o_g$, input window length $K$, planning horizon $T$, population size $N$, elite count $M_\text{elite}$, CEM iterations $J$
  \STATE \commentcode{extract the physics descriptor token of the test scene and encode observations}
  \STATE $\zp^\text{ev} \leftarrow \ctxenc(\mathcal{C}_{\text{eval}})$ ~;~ $z_0 \leftarrow E_\text{v}(o_0)$ ~;~ $z_g \leftarrow E_\text{v}(o_g)$
  \STATE \commentcode{initialize the Gaussian distribution over candidate action sequences of horizon $T$}
  \STATE $\boldsymbol{\mu} \leftarrow \mathbf{0}_{T \times d_a}$ ~;~ $\boldsymbol{\sigma} \leftarrow \mathbf{1}_{T \times d_a}$
  \For{ $j = 1 \to J$ }
    \For{ $n = 1 \to N$ }
      \STATE \commentcode{sample a candidate action sequence of length $T$ and initialize $\hat{z}_0^{(n)} = z_0$}
      \STATE $a_{0:T-1}^{(n)} \leftarrow \boldsymbol{\mu} + \boldsymbol{\sigma} \odot \epsilon^{(n)}$, $\epsilon^{(n)} \sim \mathcal{N}(\mathbf{0}, \mathbf{I})$ ~;~ $\hat{z}_0^{(n)} \leftarrow z_0$
      \For{ $t = 0 \to T-1$ }
        \STATE \commentcode{encode the $t$-th action of the $n$-th candidate action sequence to action latents}
        \STATE $e_t^{a,(n)} \leftarrow E_\text{a}(a_t^{(n)})$ 
        \STATE \commentcode{predict future latents autoregressively}
        \STATE $\hat{z}_{t+1}^{(n)} \leftarrow f_\theta\left( \{(\hat{z}_i^{(n)}, e_i^{a,(n)})\}_{i=\max(0,\, t - K + 1)}^t,\; \zp^\text{ev} \right)$ 
      \EndFor
      \STATE $C^{(n)} \leftarrow \|\hat{z}_T^{(n)} - z_g\|_2^2$ \commentcode{compute the cost of the $n$-th action sequence at CEM iteration $j$}
      \label{app:algo:cem_compute_cost}
    \EndFor

  \STATE \commentcode{select the indexes of the  $M_\text{elite}$ candidates with the lowest cost}
\STATE $\mathcal{S} \leftarrow \{\, m : C^{(m)} \text{ is among the } M_\text{elite} \text{ smallest costs} \,\}$
\STATE \commentcode{refit the action distribution using the elite set}
\STATE $\boldsymbol{\mu} \leftarrow \frac{1}{M_\text{elite}} \sum_{m \in \mathcal{S}} a^{(m)}_{0:T-1}$ ~;~ $\boldsymbol{\sigma} \leftarrow \Big(\frac{1}{M_\text{elite}} \sum_{m \in \mathcal{S}} \big(a^{(m)}_{0:T-1} - \boldsymbol{\mu}\big)^{\odot 2}\Big)^{\odot \frac{1}{2}}$
    
  \EndFor
  \STATE \textbf{return} planned action sequence $a_{0:T-1}^* \leftarrow \boldsymbol{\mu}$

\end{algorithmic}
\end{algorithm}
\vspace{-10pt}

\begin{table}[H]
    \centering
    \small
    \setlength{\tabcolsep}{10pt}
    \begin{tabular}{lcc}
        \toprule
        \textbf{Hyperparameter} & \textbf{Push-T \& Reacher} & \textbf{FetchSlide \& BeerPong} \\
        \midrule
        Population size ($N$) & 300 & 64 \\
        Elite count ($M_\text{elite}$) & 30 & 8 \\
        CEM iterations ($J$) & 30 & 8 \\
        Planned action length ($T$) & 5 model steps (25 actions) & 1 initial strike / throw command \\
        \bottomrule
    \end{tabular}
    \vspace{-7pt}
    \caption{\textbf{Hyperparameters for Visual Latent Planning with CEM.} FetchSlide and BeerPong use a smaller optimization budget since they only optimize significantly shorter action sequences.}
    \label{tab:cem_params}
\end{table}
\vspace{-5pt}

%% file: tables/table_next_frame_four_envs.tex
\begin{table}[H]
    \centering
    \small
    \setlength{\tabcolsep}{4pt}
    \adjustbox{max width=\textwidth}{%
    \begin{tabular}{l*{8}{c}}
        & \multicolumn{4}{c}{\textbf{Push-T}} & \multicolumn{4}{c}{\textbf{Reacher}} \\
        \cmidrule(lr){2-5}\cmidrule(lr){6-9}
        \textbf{Method} & PSNR$\uparrow$ & SSIM$\uparrow$ & LPIPS$\downarrow$ & IoU(\%)$\uparrow$ & PSNR$\uparrow$ & SSIM$\uparrow$ & LPIPS$\downarrow$ & IoU(\%)$\uparrow$ \\
        \midrule
        TS-WM & 24.95 & 0.975 & 0.0172 & 80.3 & 22.75 & 0.977 & 0.0188 & 57.7 \\
        DINO-WM & 27.12 & 0.981 & 0.0097 & 85.9 & 23.79 & 0.981 & 0.0125 & 65.3 \\
        ~ + Long History & 27.27 & 0.982 & 0.0094 & 86.4 & \textbf{26.78} & \textbf{0.986} & \textbf{0.0063} & \textbf{77.9} \\
        ~ + Context Prefix & 27.07 & 0.981 & 0.0096 & 86.2 & 24.62 & 0.982 & 0.0104 & 69.6 \\
        ~ + Estimated Physics$^\dagger$ & \textbf{28.58} & \textbf{0.985} & \textbf{0.0076} & \textbf{89.3} & \underline{25.21} & \underline{0.984} & \underline{0.0096} & \underline{71.6} \\
        \rowcolor{green!17}
        ~ + \shortmethod (Ours) & \underline{28.29} & \underline{0.984} & \underline{0.0080} & \underline{88.7} & 25.02 & \underline{0.984} & 0.0097 & 71.3 \\
        \midrule
        LeWM & \underline{27.50} & \underline{0.974} & \underline{0.0292} & \underline{85.8} & \underline{24.13} & \textbf{0.974} & \textbf{0.0222} & \underline{66.9} \\
        \rowcolor{green!17}
        ~ + \shortmethod (Ours) & \textbf{28.74} & \textbf{0.977} & \textbf{0.0261} & \textbf{88.4} & \textbf{24.39} & \underline{0.973} & \underline{0.0242} & \textbf{67.4} \\
        \midrule
        & \multicolumn{4}{c}{\textbf{BeerPong}} & \multicolumn{4}{c}{\textbf{FetchSlide}} \\
        \cmidrule(lr){2-5}\cmidrule(lr){6-9}
        \textbf{Method} & PSNR$\uparrow$ & SSIM$\uparrow$ & LPIPS$\downarrow$ & IoU(\%)$\uparrow$ & PSNR$\uparrow$ & SSIM$\uparrow$ & LPIPS$\downarrow$ & IoU(\%)$\uparrow$ \\
        \midrule
        TS-WM & 33.71 & 0.996 & 0.0043 & 62.5 & 29.15 & 0.990 & 0.0071 & 84.7 \\
        DINO-WM & 35.81 & \underline{0.997} & 0.0026 & 73.7 & 34.35 & \underline{0.996} & 0.0021 & 93.4 \\
        ~ + Long History & \textbf{36.98} & \textbf{0.998} & \textbf{0.0018} & \textbf{78.6} & \textbf{35.49} & \textbf{0.997} & \textbf{0.0016} & \underline{94.4} \\
        ~ + Context Prefix & 35.88 & \underline{0.997} & 0.0026 & 74.8 & 34.73 & \underline{0.996} & \underline{0.0017} & 93.7 \\
        ~ + Estimated Physics$^\dagger$ & \underline{36.50} & \textbf{0.998} & 0.0022 & 76.5 & 34.57 & \underline{0.996} & 0.0018 & 93.4 \\
        \rowcolor{green!17}
        ~ + \shortmethod (Ours) & 36.42 & \textbf{0.998} & \underline{0.0021} & \underline{76.6} & \underline{35.45} & \textbf{0.997} & 0.0018 & \textbf{94.7} \\
        \midrule
        LeWM$^*$ & \underline{30.31} & \underline{0.987} & \underline{0.0153} & \underline{67.2} & \underline{29.77} & \textbf{0.986} & \textbf{0.0283} & \underline{78.6} \\
        \rowcolor{green!17}
        ~ + \shortmethod (Ours) & \textbf{31.61} & \textbf{0.989} & \textbf{0.0140} & \textbf{72.8} & \textbf{29.94} & \underline{0.985} & \underline{0.0286} & \textbf{79.3} \\
        \bottomrule
    \end{tabular}}
    \caption{\textbf{Quantitative Next-frame Prediction Results in Four Environments.} We report PSNR, SSIM, LPIPS, and foreground IoU (\%).
    LeWM$^*$: we use TC-LeWM in BeerPong to enhance performance (see Appendix~\ref{sec:appendix_baselines} for details). Bold and underlined values indicate the best and second-best results within each group separately: the six-method DINO-WM/TS-WM group and the two-method LeWM group.}
    \label{tab:next_frame_four_envs}
\end{table}